\documentclass[letterpaper, 10 pt, conference]{ieeeconf}  

\IEEEoverridecommandlockouts                              

\usepackage{graphics} 
\usepackage{epsfig} 
\usepackage{subfigure}
\usepackage{amsmath} 
\usepackage{epstopdf}
\usepackage{lipsum,amsmath,multicol}
\usepackage{float}
\usepackage{algorithm}
\usepackage{algorithmicx}
\usepackage{algpseudocode}
\usepackage{tcolorbox}

\newtheorem{theorem}{Theorem}

\title{\LARGE \bf
Combining Method of Alternating Projections and Augmented Lagrangian for Task Constrained Trajectory Optimization.
}

\author{Arun Kumar Singh$^{1}$, Reza Ghabcheloo $^{1}$, Andreas Muller$^{2}$, Harit Pandya${^3}$
\thanks{The research was partly supported by TUT postdoc fellowship to the first author. $^{1}$ Department of Hydraulics and Automation, Tampere University of Technology, Finland. ${^2}$ Johannes Kepler University, Austria, ${^3}$ Robotics Research Center, IIIT-Hyderabad, India}
}

\begin{document}

\maketitle
\thispagestyle{empty}
\pagestyle{empty}

\allowdisplaybreaks

\begin{abstract}
Motion planning for manipulators under task space constraints is difficult as it constrains the joint configurations to always lie on an implicitly defined manifold. It is possible to view task constrained motion planning as an optimization problem with non-linear equality constraints which can be solved by general non-linear optimization techniques. In this paper, we present a novel custom  optimizer which exploits the underlying structure present in many task constraints. 

At the core of our approach are some simple reformulations, which when coupled with the \emph{method of alternating projection}, leads to an efficient convex optimization based routine for computing a feasible solution to the task constraints.  We subsequently build on this result and use the concept of Augmented Lagrangian  to guide the feasible solutions towards those which also minimize the user defined cost function. We show that the proposed optimizer is fully distributive and thus, can be easily parallelized. We validate our formulation on some common robotic benchmark problems.  In particular, we  show that the proposed optimizer achieves cyclic motion in the joint space corresponding to a similar nature trajectory in the task space. Furthermore, as a baseline, we compare the proposed optimizer with an off-the-shelf non-linear solver provide in open source package SciPy. We show that for similar task constraint residuals and smoothness cost, it can be upto more than three times faster than the SciPy alternative.

\end{abstract}

\section{Introduction}

Presence of task or kinematic constraints on end-effector orientation and position dramatically increases the difficulty of the motion planning problem.  As explained in \cite{berenson}, \cite{kino_manifold}, the reason for this could be traced to the fact that task constraints gives rise to an implicit manifold on which the joint configurations must always lie. In addition, directly sampling from this implicit manifold is extremely difficult. In this paper, we view  the problem of task constrained motion planning through the lens of trajectory optimization. In this context, the computational bottleneck exists due to the non-linear equalities arising out of task constraints. Although, there exists various optimization  techniques like sequential quadratic programming (SQP) and Newton's method to locally solve such non-linear and non-convex optimization problems, they are not  designed to specifically consider the hidden underlying structure present in many task constraints, exploiting which could lead to significant computational advantage. Customization of optimization routines has been actively pursued in robotics motion planning \cite{chomp}, \cite{trajopt} and the current proposed work contributes to this line of research.

One of the important challenges in mathematical optimization is to derive globally valid convex approximation{\footnote{Here, globally valid means that the convex approximation of the original non-convex problem holds everywhere in the variable space. This is unlike affine approximation used in SQP obtained through Taylor series expansion which is only valid in the vicinity of the expansion point.}} for non-convex problems as they can be solved without the additional computational bottleneck of incorporating trust region constraints{\footnote{Non-linear optimization techniques like SQP require  additional so called trust region constraints which forces the solution at each iteration to be in the vicinity of that obtained in the previous iteration.  The size of the trust region influences the progress of the optimization. However, there are no established techniques to a priori estimate the size of the trust region}}. One such popular convex approximation called \emph{convex-concave procedure} (CCP) \cite{ccp1} has already proved immensely useful in robotics for a diverse set of  motion planning problems \cite{ccp_robot1}, \cite{ccp_robot2}. However, the highly non-linear nature of manipulator kinematics and task constraints precludes the use of CCP for task constrained trajectory optimization.

\noindent \textbf{Contributions and Main Results}:On the theoretical side, we present a new globally valid convex approximation for the problem of task constrained trajectory optimization. The foundation of our approach is built on some simple reformulations which, when coupled with \emph{method of alternating projection} (MAP) \cite{map1}, \cite{map2}, \cite{map3}, \cite{map4}, leads to a convex optimization problem for computing a feasible solution to the task constraints. As a natural next step, we use the concept of Augmented Lagrangian (AL) \cite{al1}, \cite{al2} to  guide the MAP towards solutions which are not only feasible but also minimize a user defined cost function. An added feature of the proposed optimization is that it is fully distributive and thus can be easily parallelized across multiple processors. 

On the implementation side, we consider the following applications: (i) Computing smooth joint motions to execute the given position and orientation trajectory of the end effector. We specifically consider closed cyclic trajectories in the end-effector position space and show that the corresponding joint trajectories are also cyclic and closed. This solves the well known \emph{cyclicity} bottleneck  associated with task space planning of redundant manipulators \cite{kuka_cyclic}. In \cite{kuka_cyclic}, authors proposed a randomized sampling based solution. In contrast, we present an optimization based approach. (ii) Computing smooth joint and task space trajectory for point to point motion with trajectory wide constraints on end effector orientation and/or partial constraints on positions. (iii) Computing smooth joint and task space trajectory for point to point motion of a redundantly actuated closed kinematic chain.


\section{Preliminaries}\label{prelim}

\noindent \textbf{Augmented Lagrangian (AL)}:Consider, the following optimization with convex cost and affine equality constraints in terms of variable $\textbf{v}$.

\small
\begin{equation}
\min f(\textbf{v}) \hspace{0.1cm}\text{such that} \hspace{0.1cm} \textbf{S}\textbf{v}=\textbf{r}
\end{equation}
\normalsize

\noindent The AL technique solves this problem by incorporating the affine equality constraints as a penalty in the cost function \cite{al1}, \cite{al2}.

\small
\begin{equation}
f(\textbf{v})+\overbrace{\boldsymbol{\lambda}^T(\textbf{S}\textbf{v}-\textbf{r})+\rho \Vert \textbf{S}\textbf{v}-\textbf{r} \Vert_2^2}^{augmented \hspace{0.2cm} lagrangian}.
\label{auglagrange}
\end{equation}
\normalsize

\noindent Where, $\boldsymbol{\lambda} $ is called the Lagrange multipliers and $\rho$ is a positive constant which we will  henceforth call as proximal weights (since they are associated with the proximal operator). The solution of (\ref{auglagrange}) can be computed through the following iterates. In the following, $\Delta$ is a constant and further, $\Delta\geq 1$.

\vspace{-0.5cm}

\small
\begin{subequations}
\begin{align*}
\textbf{v}^{k+1} = \arg\min (f(\textbf{v})+ (\boldsymbol{\lambda}^k)^T(\textbf{S}\textbf{v}-\textbf{r})+\rho^k\Vert \textbf{S}\textbf{v}-\textbf{r} \Vert_2^2)\\
\boldsymbol{\lambda}^{k+1} = \boldsymbol{\lambda}^k + \rho^k(\textbf{S}\textbf{v}-\textbf{r})\\
\rho^{k+1} = \rho^k\Delta
\end{align*}
\end{subequations}
\normalsize


\noindent\textbf{Method of Alternating Projection (MAP)}: Given two sets $\mathcal{C}$, $\mathcal{D}$, the \emph{method of alternating projection} or MAP computes an  intersection point between the two sets. Formally, the problem can be framed as the following minimization.
\small
\begin{equation}
\textbf{v}, \textbf{w} = \arg\min_{\textbf{v}\in\mathcal{C},\textbf{w}\in\mathcal{D}} \Vert\textbf{v}-\textbf{w}\Vert_1
\label{proj_min}
\end{equation} 
\normalsize

\noindent The minimization (\ref{proj_min}) proceeds through the following iterates.

\small
\begin{equation}
\textbf{v}^{k+1} = \arg\min_{\textbf{v}\in \mathcal{C}} \Vert\textbf{v}-\textbf{w}^k\Vert_1, \textbf{w}^{k+1} = \arg\min_{\textbf{w}\in \mathcal{D}} \Vert\textbf{w}-\textbf{v}^{k+1}\Vert_1
\label{step_map}
\end{equation}
\normalsize


\noindent The first step in (\ref{step_map}) projects $\textbf{w}^{k}$ to set $\mathcal{C}$ and subsequently, the second step  projects $\textbf{v}^{k+1}$ to set $\mathcal{D}$. Although initially proposed for convex $\mathcal{C}$, $\mathcal{D}$, MAP has been shown to work well for non-convex sets as well \cite{map2}, \cite{map3}, \cite{map4}.

\noindent \textbf{Alternating Minimization (AM):} Alternating minimization (AM) or Gauss-Seidel method can be seen as a generalization of MAP  to include arbitrary cost functions instead of just distance between two points. Given a function, $f(\textbf{v},\textbf{w})$, AM proceeds through the following iterates \cite{boyd_scp}.

\small
\begin{equation}
\textbf{v}^{k+1} =\arg\min f(\textbf{v}, \textbf{w}^{k}),\textbf{w}^{k+1} =\arg\min f(\textbf{w}, \textbf{v}^{k+1})
\end{equation}
\normalsize

\noindent AM proves extremely useful when $f(\textbf{v}, \textbf{w})$ is convex in $\textbf{v}$ for a given $\textbf{w}$ and vice versa but is non-convex when $\textbf{v}$ and $\textbf{w}$ are jointly considered.

\section{Problem Formulation}

\subsection{Task Constraints}

\noindent We consider task constraints which appear as kinematic constraints on end-effector's position, orientation or both. For position level constraints, we assume having access to vector $\textbf{x}_t^{des}=(x_t^{des}, y_t^{des}, z_t^{des})$, which models the desired position at time $t$. Similarly, for the orientation constraints, we assume that we have access to  a desired rotation matrix $\textbf{R}_t^{des}$ at time $t$ which in turn is used to construct a  vector $\textbf{r}_t^{des}$ formed with the elements of the rotation matrix. Thus, its dimension, $ndim(\textbf{r}_t^{des})=9, \forall t$. In contrast, we allow for flexibility in $ndim(\textbf{x}_t^{des})$ to account for the  fact that sometimes we have only partial constraints on the end-effector position restricting motion only along some directions.

\subsection{Task Constrained Trajectory Optimization}
\noindent We consider trajectories in fixed time interval $[0, t_f]$ which in turn is discretized into $n$ grid points. Let, $\textbf{q}_t$ represent the configuration at time $t$. Further, each configuration is supposed to be composed of $m$ joint angles. That is, $\textbf{q}_t=(q_t^1, q_t^2...q_t^m)$. Stacking all the configurations at different time instants, we obtain a matrix $\textbf{Q}=(\textbf{q}_{t_0}, \textbf{q}_{t_2},... \textbf{q}_{t_n})$ representing the joint trajectory in $[0, t_f]$. For a kinematic robot control problem, the joint velocities can be taken as the control input. Thus, $\frac{\textbf{q}_{t+1}-\textbf{q}_{t}}{\Delta t}$ can be considered as the control input which drives the joint from configuration $\textbf{q}_{t}$ to $\textbf{q}_{t+1}$ in time duration, $\Delta t$. With these notations, in place, we follow the construction of \cite{toussiant} and frame the task constrained trajectory optimization in the following manner.

\small
\begin{subequations}
\begin{align}
\arg\min_{\textbf{Q}} \sum_{t=0}^{t=t_f} J(\textbf{q}_{t-k:t}) \label{cost}\\
\textbf{q}_{t} \in \mathcal{C}_{\textbf{q}} \label{feasible_set}\\
\textbf{f}_{t}(\textbf{q} ) = ({^1}{f}_{t}(\textbf{q}_{t}), {^2}{f}_{t}(\textbf{q}_{t})...{^d}{f}_{t}(\textbf{q}_{t} )) = 0 \label{task_const}
\end{align}
\end{subequations}
\normalsize

\noindent where, $J(\textbf{q}_{t-k:t})$ represents a user defined convex cost function which depends on all the configurations in time interval $[t-k, \hspace{0.1cm} t]$. For example, choosing $k=2$ we can derive an approximate sum of squared of  accelerations, (\ref{sum_acc}). 
\small
\begin{equation}
J(\textbf{q}_{t-2:t}) = \Vert \textbf{q}_{t-2}+\textbf{q}_{t}-2\textbf{q}_{t-1}\Vert_2^2
\label{sum_acc}
\end{equation}
\normalsize

Throughout this paper, we use (\ref{sum_acc}) as our default cost function. However, other choices can be easily incorporated within the optimization. For example, choosing $k=1$, one can derive the sum of square velocity cost.

The equation (\ref{feasible_set}) constrain the joint configurations to lie in the feasible set, $\mathcal{C}_{\textbf{q}}$ which in our case models the  joint limits. Thus, $\mathcal{C}_{\textbf{q}} = [-\textbf{q}_{max} \hspace{0.2cm} \textbf{q}_{max}]$. We do not include any velocity or acceleration constraints into the definition of feasible set as these can be satisfied through trajectory re-timing based on time scaling \cite{cuong_topp}, \cite{aks_topp}. The non-linear equalities, (\ref{task_const}) models the  task space constraints at time  $t$. Further, each function, $\textbf{f}_t(.)$ has $d$ components depending on whether the task constraints include position, orientation or both. Further, $d$ also depends on the $ndim(\textbf{x}_t^{des})$.

\subsection{Algebraic Representation of Task Constraints}
\noindent Here, we derive a niche representation for each component of task constraints $\textbf{f}_t(.)$ which later on forms the basis of our proposed optimizer. The representation is derived from the following theorem.

\noindent \begin{theorem}
Each ${^d}f_t(\textbf{q})$ can always be represented in the following form.
\small
\begin{equation}
  {^d}f_t(\textbf{q}_t)=\begin{cases}
    {^d}g_t^{1}(.)\cos(q^1_t)+{^d}h_t^{1}()\sin(q^1_t)+{^d}p_t^{1}(.)\\
    {^d}g_t^2(.)\cos(q_t^2)+{^d}h_t^2(.)\sin(q_t^2)+{^d}p_t^2(.)\\
...................................\\
{^d}g_t^m(.)\cos(q_t^m)+{^d}h_t^m(.)\sin(q_t^m)+{^d}p_t^m(.)\\    
  \end{cases}
  \label{nonlin_gen_spatial}
\end{equation}
\normalsize

\noindent Where, ${^d}g_t^j(.)$, ${^d}h_t^j(.)$ and ${^d}p_t^j(.)$ are functions which may depend on sine and cosine of all the joint angles except $q_t^j$. For example, ${^d}g_t^1(.)$ may depend on $(\cos(q_t^2), \sin(q_t^2))$,  $(\cos(q_t^3), \sin(q_t^3))$...$(\cos(q_t^m), \sin(q_t^m))$ but would be completely independent of $q_t^1$.
\end{theorem}

Due to space restrictions, we do not present the exact proof but point out that it draws from the separable nature of forward kinematics of manipulators where the end transformation is a product of individual transformation matrices. Consequently, a slight algebraic manipulation will lead us to (\ref{nonlin_gen_spatial}).

\noindent  A simpler special case of (\ref{nonlin_gen_spatial}) can be written in the following manner:

\small
\begin{equation}
{^d}{f}_{t}(\textbf{q}_t) = \sum_{j=1}^{j=m} {^d} a_{j} \cos(\textbf{A}_j\textbf{q}_{t})+{^d}b_{j}\sin(\textbf{A}_j\textbf{q}_{t})+{^d}c_{j},
\label{nonlin_gen}
\end{equation}
\normalsize

\noindent Where, ${^d}a_{j}, {^d}b_{j}, {^d}c_{j}$ are scalar constants while each $\textbf{A}_j$ is a constant row matrix with number of columns equal to  $ndim(\textbf{q}_t)$.

\section{Main Result}
In this section, we present the main theoretical result of this paper;  a custom optimizer for the problem (\ref{cost})-(\ref{task_const}). We first derive the optimizer for the simpler task constraints (\ref{nonlin_gen}) and then subsequently extend it to accommodate the more general variant (\ref{nonlin_gen_spatial}). The derivation proceeds by first formulating an MAP update for a feasibility variant of  (\ref{cost})-(\ref{task_const}) which is subsequently extended to the general optimization case.

\subsection{Proposed Optimizer with task constraints, (\ref{nonlin_gen})}

\noindent We begin by introducing the  slack variables   $v_t^1, v_t^2....v_t^m$ and $w_t^1, w_t^2....w_t^m$, $\vert v_t^j\vert \leq 1$, $\vert w_t^j\vert \leq 1$ at time  $t$. These slack variables can be related  to the task constraints through the following relations:

\small
\begin{equation}
v_{t}^j = \cos(\textbf{A}_j\textbf{q}_i), w_{t}^j = \sin(\textbf{A}_j\textbf{q}_i) \forall j = 1,2..m
\label{transform} 
\end{equation}
\normalsize

\noindent Using (\ref{transform}), the task constraints (\ref{nonlin_gen}) gets reformulated in the following manner. Note how, (\ref{nonlin_gen2}) is affine with respect to $v_t^j$ and $w_t^j$.

\small
\begin{eqnarray}
{^d}f_t(v_t^j, w_t^j ) =  \sum_{j=0}^{j=m} {^d}a_{j} v_t^j+ {^d}b_{j} w_t^j+{^d}c_j=0 \label{nonlin_gen2}
\end{eqnarray}
\normalsize

\noindent Using (\ref{nonlin_gen2}), we formulate the following  feasibility problem formed with the constraints of optimization (\ref{cost})-(\ref{task_const}):

\small
\begin{subequations}
\begin{align}
Solve, \textbf{q}_t \in \mathcal{C} \label{feasible_set2}\\
\textbf{f}_{t}(\textbf{v}_t, \textbf{w}_t ) =0,\forall t\in [0 \hspace{0.2cm} t_f] \label{task_const2}\\
\textbf{v}_{t} = \textbf{b}_{v_{t}}(\textbf{q}_{t}) ,\forall t\in [0 \hspace{0.2cm} t_f]  \label{slack1}\\
\textbf{w}_{t} = \textbf{b}_{w_{t}}(\textbf{q}_{t}),\forall t\in [0 \hspace{0.2cm} t_f] \label{slack2}\\
\vert \textbf{v}_t\vert \leq 1, \vert \textbf{w}_t\vert \leq 1\label{vw_bound}
\end{align}
\end{subequations}
\normalsize

\noindent In (\ref{slack1})-(\ref{slack2}), the vectors, $\textbf{v}_{t}$, $\textbf{w}_{t}$, $\textbf{b}_{v_{t}}(\textbf{q}_{t})$ and $\textbf{b}_{w_{t_i}}(\textbf{q}_{t_i})$ as defined in (\ref{vw_vec})-(\ref{cossin_vec})  are used to put (\ref{transform}) in matrix form. Also, note that the inequality, (\ref{vw_bound}) holds component-wise. 

\small
\begin{equation}
\textbf{v}_{t} = \begin{bmatrix}
v_{t}^1\\
v_{t}^2\\
.\\
.\\
v_{t}^m
\end{bmatrix}, \textbf{w}_{t} = \begin{bmatrix}
w_{t}^1\\
w_{t}^2\\
.\\
.\\
w_{t}^m
\end{bmatrix}
\label{vw_vec}
\end{equation}
\normalsize

\small
\begin{equation}
\textbf{b}_{\textbf{v}_{t}}(\textbf{q}_{t}) = \begin{bmatrix}
\cos(\textbf{A}_1\textbf{q}_{t})\\
\cos(\textbf{A}_2\textbf{q}_{t})\\
.\\
.\\
\cos(\textbf{A}_m\textbf{q}_{t})
\end{bmatrix}, \textbf{b}_{\textbf{w}_{t}}(\textbf{q}_{t}) = \begin{bmatrix}
\sin(\textbf{A}_1\textbf{q}_{t})\\
\sin(\textbf{A}_2\textbf{q}_{t})\\
.\\
.\\
\sin(\textbf{A}_m\textbf{q}_{t})
\end{bmatrix}
\label{cossin_vec}
\end{equation}
\normalsize

\noindent The solution to the feasibility program boils down to computing the intersection of the following  two sets at each time instant $t$.

\small
\begin{subequations}
\begin{align}
\mathcal{C}_{t} = \{(\textbf{v}_{t},\textbf{w}_{t}): \textbf{f}_{t}(v_{t}^j, w_{t}^j ) = 0 \} \label{set_ct}\\
\mathcal{D}_{t} = \{ (\cos(\textbf{A}_j\textbf{q}_{t}), \sin(\textbf{A}_j\textbf{q}_{t})), \forall j : \textbf{q}_{t}\in \mathcal{C}_{\textbf{q}} \}\label{set_dt}
\end{align}
\end{subequations}
\normalsize

\noindent As evident, the set $\mathcal{C}_{t}$ contains all the pairs of $(\textbf{v}_t,\textbf{w}_t^j)$ which satisfy the task constraints. Similarly, $\mathcal{D}_{t}$ contains all the pairs of $(\cos(\textbf{A}_j\textbf{q}_{t}), \sin(\textbf{A}_j\textbf{q}_{t}))$ for each feasible joint configuration, $\textbf{q}_{t}$. We use MAP for computing the intersection point of these two sets. Starting with a point, $\textbf{q}_{t}^k$, Algorithm \ref{algo1} provides the MAP updates for computing a feasible solution. The projection onto set $\mathcal{C}_{t}$ is shown on line 2 and is self explanatory. The minimization on line 5 performs the projection onto the set $\mathcal{D}_{t}$ and to see how, note the following reformulation, where $A$ is defined in Algorithm \ref{algo1}.

\small
\begin{eqnarray}\nonumber
\arg\min_{\textbf{q}_{t}}( \Vert (\cos(\textbf{A}\textbf{q}_{t})-\textbf{v}_{t}\Vert_1+\Vert (\sin(\textbf{A}\textbf{q}_{t})-\textbf{w}_{t}\Vert_1) \nonumber\\
\Rightarrow \arg\min \Vert  \textbf{A}\textbf{q}_{t}-\arctan2(\frac{\textbf{w}_{t}}{\textbf{v}_{t}}) \Vert_1\nonumber
\end{eqnarray}
\normalsize

\small
\begin{algorithm}
 \caption{MAP update for a Feasible Solution to Task Constraints}\label{algo1}
    \begin{algorithmic}[1]
    \small
\State Define \begin{equation*}
\textbf{A} = \begin{bmatrix}
\textbf{A}_1 & \textbf{A}_2 &....&\textbf{A}_m\\
\end{bmatrix}^T
\end{equation*}

     \While{$k\leq maxiter$}
           \For{$t= 0:t_f$}
           \small
          \begin{eqnarray}
(\textbf{v}_{t}^{k+1},\textbf{w}_{t}^{k+1}) = \arg\min \Vert \textbf{v}_{t}- \textbf{b}_{v_{t}}(\textbf{q}^k_{t}))\Vert_1 \nonumber \\+ \Vert \textbf{w}_{t}- \textbf{b}_{w_{t}}(\textbf{q}^k_{t}))\Vert_1 \nonumber\\
\text{such that},\textbf{f}_{t}( \textbf{v}_t, \textbf{w}_t )=0, \vert \textbf{v}_t\vert \leq 1, \vert \textbf{w}_t\vert \leq 1\label{first_proj}
\end{eqnarray}
\normalsize
\EndFor

\For{$t= 0:t_f$}
\small
\begin{eqnarray}
\textbf{q}_t^{k+1} = \arg\min \Vert  \textbf{A}\textbf{q}_{t}-\arctan2(\frac{\textbf{w}_{t}^{k+1}}{\textbf{v}_{t}^{k+1}}) \Vert_1 \nonumber\\
\text{such that}, \textbf{q}_t\in \mathcal{C}_{\textbf{q}}\label{second_proj}
\end{eqnarray} 
\normalsize
\EndFor
\EndWhile 
\normalsize          
        \end{algorithmic}  
        \end{algorithm}
\normalsize

 Two key points regarding the update rules (\ref{first_proj})-(\ref{second_proj}) are worth pointing out. Firstly, note that both the updates involves solving a convex optimization problem, although non-smooth due to the presence of $l_1$ norm. Secondly, the process of obtaining the $(\textbf{v}_{t}, \textbf{w}_{t})$ corresponding to the task constraints at time $t$ is completely independent of the  process for obtaining $(\textbf{v}_{t+1}, \textbf{w}_{t+1})$ corresponding to the task constraints at time $t+1$. Thus, the for loop  on line 2 of Algorithm \ref{algo1} can be easily parallelized across $n$ different processors. A similar argument can be made for optimization (\ref{second_proj}) where each $\textbf{q}_{t}^{k+1}$ can be obtained in parallel.

\noindent \textbf{Example}: To demonstrate the working of MAP update presented in Algorithm \ref{algo1}, we take a simple equality constraint depending on only one joint angle at time $t$
\begin{equation*}
\sin(q_t^1)+\cos(q_t^1) = 0.366, \vert {q}_t^1\vert\leq 1.0
\label{simple_task}
\end{equation*}

\begin{figure*}[!h]
  \centering
   \subfigure[]{
    \includegraphics[width= 4.35cm, height=3.6cm] {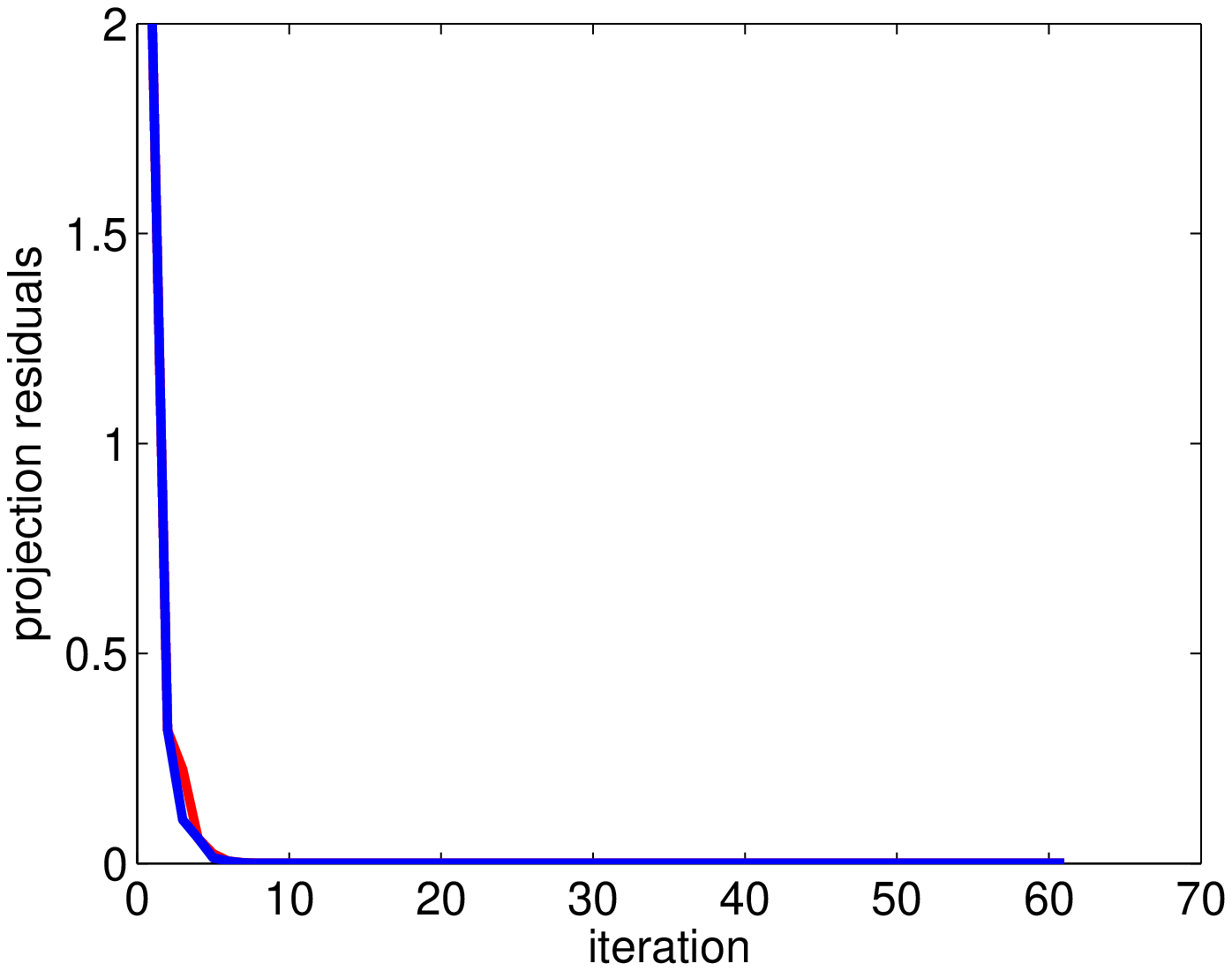}
    \label{proj_residual1}
   }\hspace{-0.7cm}
   \subfigure[]{
    \includegraphics[width= 4.35cm, height=3.6cm] {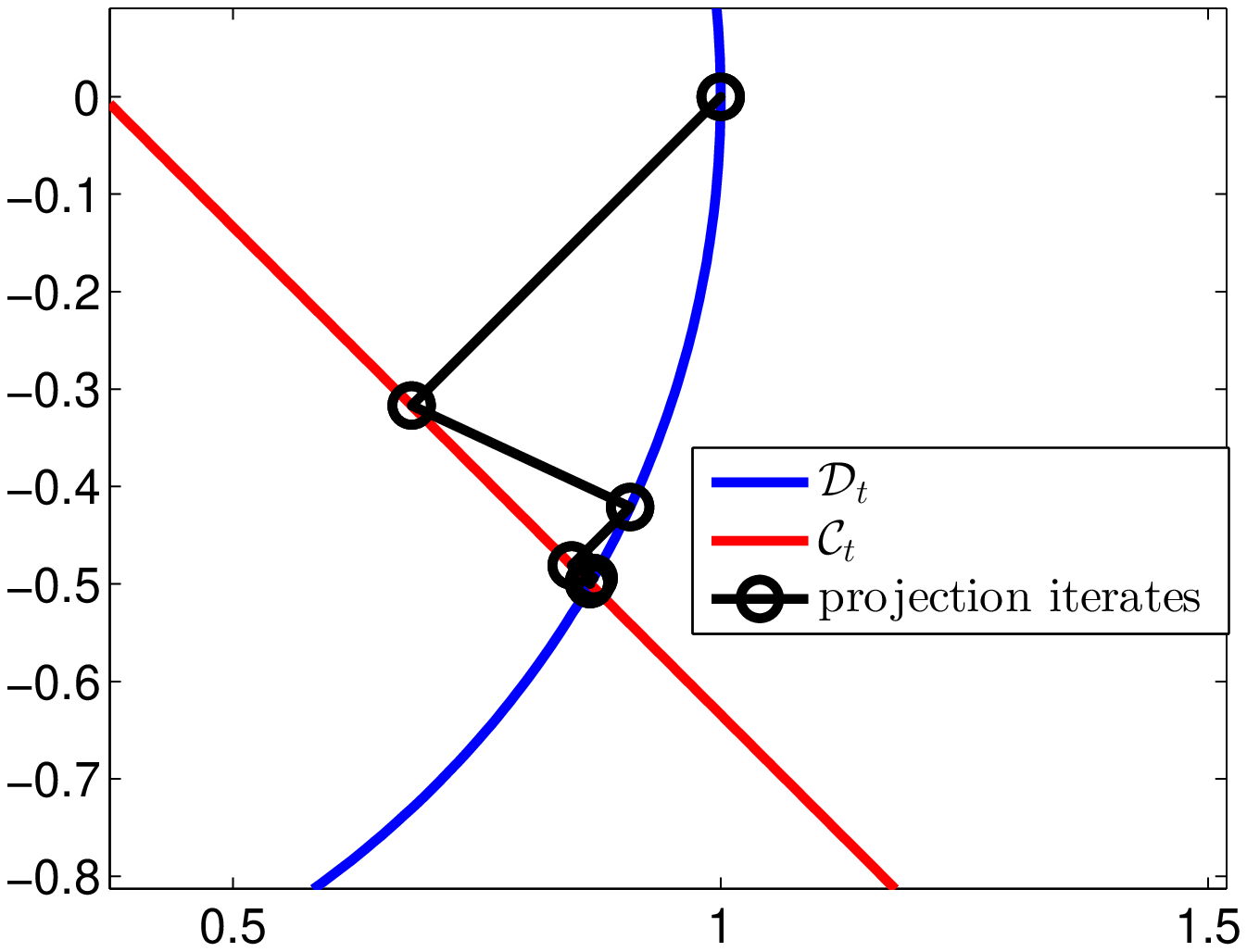}
    \label{proj_iter1}
   }\hspace{-0.7cm}
   \subfigure[]{
      \includegraphics[width= 4.35cm, height=3.6cm] {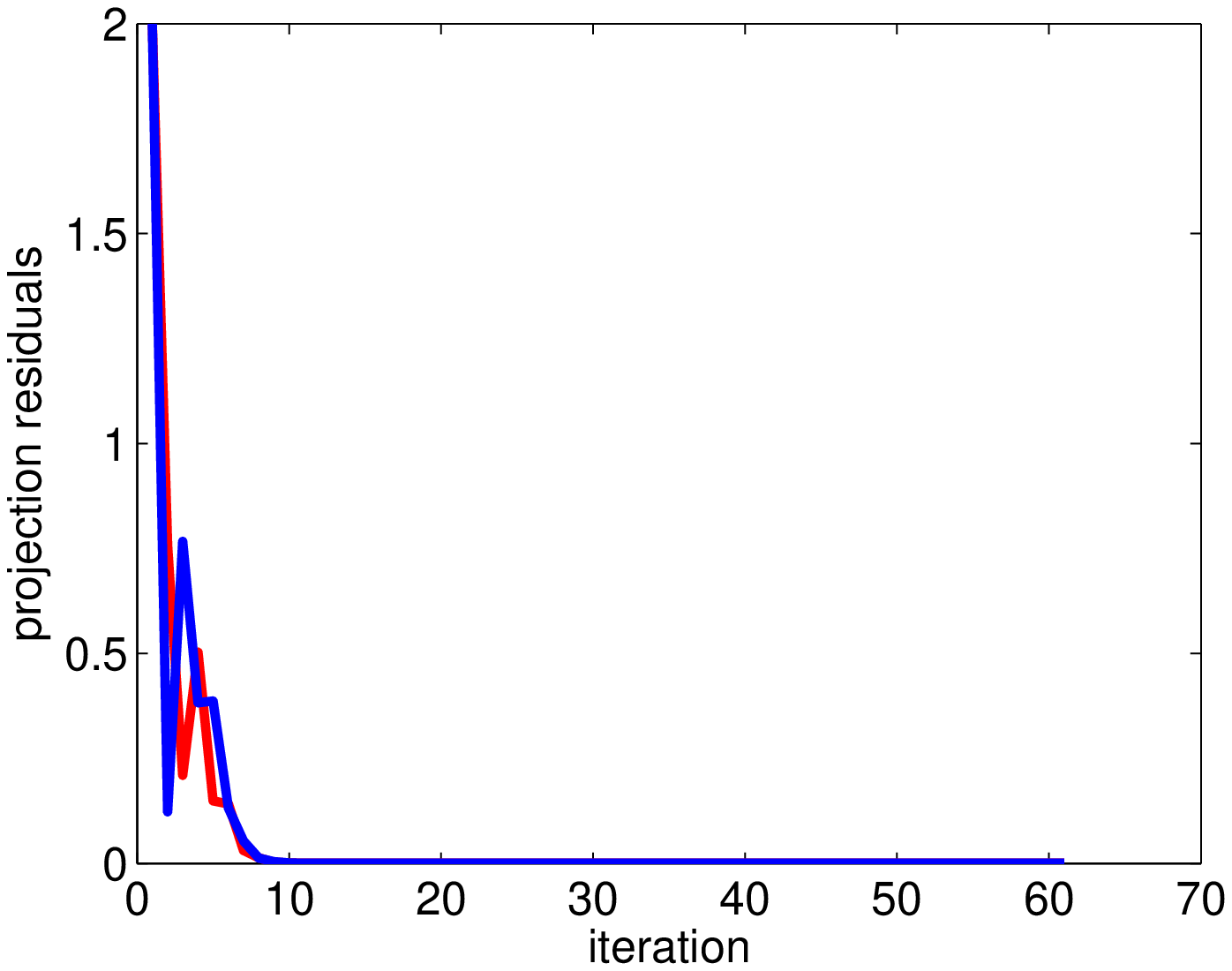}
    \label{proj_residual2}
   }\hspace{-0.5cm}
   \subfigure[]{
      \includegraphics[width= 4.35cm, height=3.6cm] {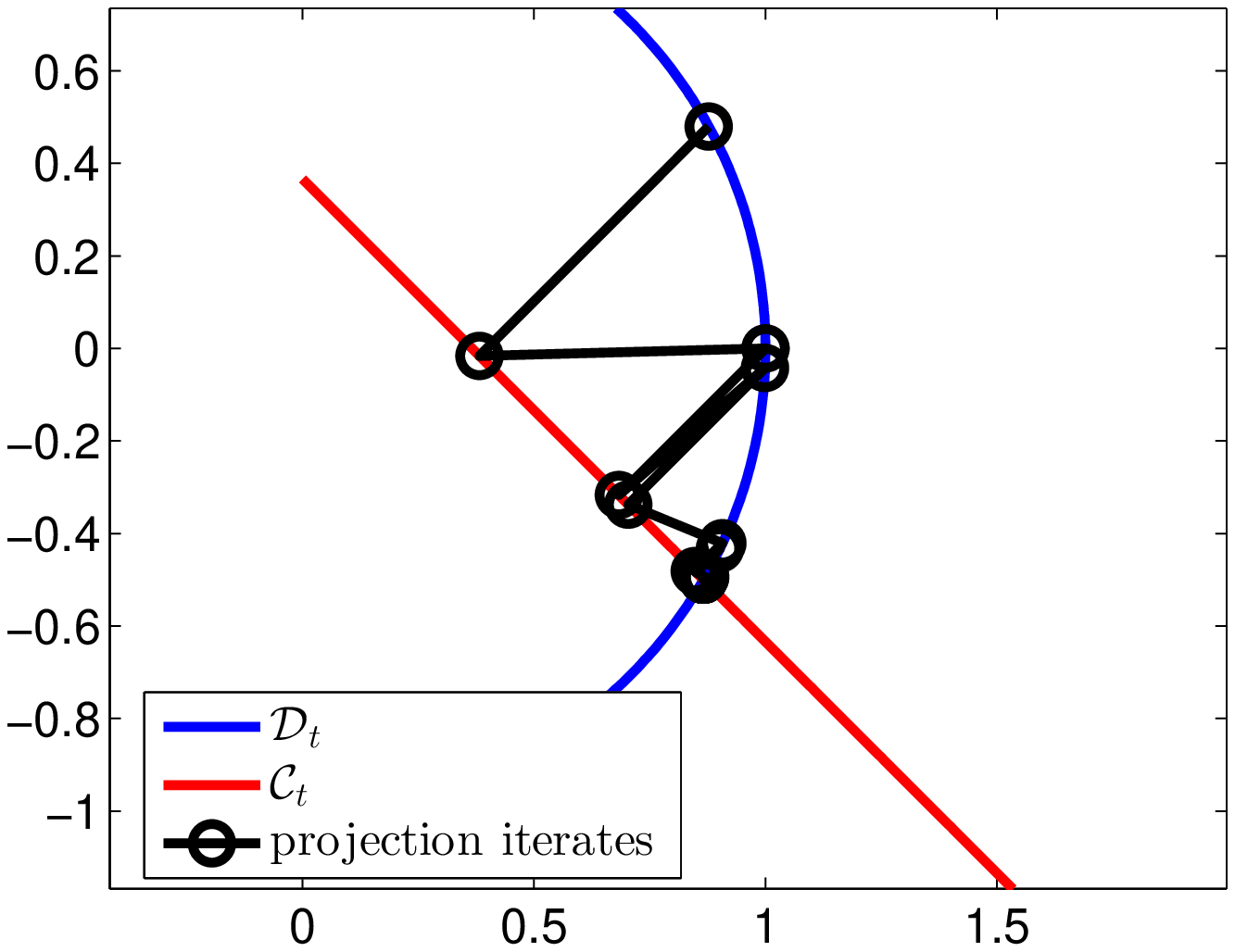}
    \label{proj_iter2}
   }
  \caption{Performance of the MAP update for a simple task constraint, (\ref{simple_task}) for two different initial guesses. Figures (a) and (c) shows the progression of the projection residuals while (b) and (d) shows the projection lines and the graphical representation of the sets, $\mathcal{C}_t$, $\mathcal{D}_t$.  }     
\end{figure*}

\noindent Using the reformulation (\ref{transform}), we introduce two new variables, $v_t^1=\cos(q_t^1), w_t^1=\sin(q_t^1)$ and consequently construct the sets (\ref{set_ct}) and (\ref{set_dt}). As shown in Fig.\ref{proj_iter1}-\ref{proj_iter2}, the sets, $\mathcal{C}_t$ and $\mathcal{D}_t$ have the topology of a straight line and a circle respectively. Fig.\ref{proj_residual1}-\ref{proj_residual2} shows the progression of the projection residual for two different initial guess of $q_t^1$.

\subsubsection{ Guiding Feasible Solutions towards Optimality}

\noindent We are now in a position to build upon  (\ref{first_proj})-(\ref{second_proj}) and derive an update rule for solving the optimization problem (\ref{cost})-(\ref{task_const}). At an intuitive level, the extension boils down to creating some sort of  mechanism to guide the feasible solution that Algorithm \ref{algo1} provides, towards those which also minimize the cost function (\ref{sum_acc}). To this end, consider the following augmented Lagrangian function with multiplier, $\boldsymbol{\lambda}_{\textbf{q}_{t}}$ and proximal weight $\rho_{\textbf{q}_{t}}$.

\small
\begin{eqnarray}
L_1 (\textbf{q}_{t} ) =   J(\textbf{q}_{t-2:t})+\boldsymbol{\lambda}_{\textbf{q}_{t}}^T(\textbf{A}\textbf{q}_{t}-\arctan2(\frac{\textbf{w}_{t}}{\textbf{v}_{t}}))\nonumber \\ 
+\rho_{\textbf{q}_{t}}\Vert\textbf{A}\textbf{q}_{t}-\arctan2(\frac{\textbf{w}_{t}}{\textbf{v}_{t}})\Vert_2^2 \label{aug1}
\end{eqnarray}
\normalsize

\noindent In (\ref{aug1}), for a give pair of $(\textbf{v}_{t}, \textbf{w}_{t})$, the terms, $\boldsymbol{\lambda}_{\textbf{q}_{t}}$ and $\rho_{\textbf{q}_{t}}$ in conjunction
controls the residual of $\Vert  \textbf{A}\textbf{q}_{t}-\arctan2(\frac{\textbf{w}_{t}}{\textbf{v}_{t}})\Vert_1$ and consequently the projection of $(\textbf{v}_{t}, \textbf{w}_{t})$ to the set of feasible $\textbf{q}_t$. Importantly, the magnitude of $\rho_{\textbf{q}_{t}}$ balances the trade-off between minimizing the cost function, $J(.)$ and the projection residual. If we always fix  $\rho_{\textbf{q}_{t}}$ to a large value, the minimization of (\ref{aug1}) and (\ref{second_proj}) becomes almost identical because in such a case, the minimization of $L_1$ would almost exclusively focus on minimizing the  projection residual.
 
 Along similar lines as above, we use AL to construct the following reformulation of (\ref{first_proj}) where, for a given $\textbf{q}_{t}$, the multipliers, $\boldsymbol{\lambda}_{\textbf{v}_{t}}, \boldsymbol{\lambda}_{\textbf{w}_{t}}, \boldsymbol{\lambda}_{\textbf{f}_{t}}$ and proximal weights, $\rho_{\textbf{v}_{t}}, \rho_{\textbf{w}_{t}}, \rho_{\textbf{f}_{t}}$ controls the  residual of the projections as wells as  that of the task constraints. Note, how in contrast to $l_1$ norm, the AL function is smooth.

\small
\begin{eqnarray}
L_2(\textbf{v}_{t}, \textbf{w}_{t} ) = \boldsymbol{\lambda}_{\textbf{v}_{t}}^T(\textbf{v}_{t}- \textbf{b}_{v_{t}}(\textbf{q}_{t}))+\rho_{\textbf{v}_{t}}\Vert \textbf{v}_{t}- \textbf{b}_{v_{t}}(\textbf{q}_{t})\Vert_2^2\nonumber \\
+\boldsymbol{\lambda}_{\textbf{w}_{t}}^T(\textbf{w}_{t}- \textbf{b}_{w_{t}}(\textbf{q}_{t}))+\rho_{\textbf{w}_{t}}\Vert \textbf{w}_{t}- \textbf{b}_{w_{t}}(\textbf{q}_{t})\Vert_2^2\nonumber\\
+\boldsymbol{\lambda}_{\textbf{f}_{t}}^T(\textbf{f}_{t}(\textbf{v}_{t},\textbf{w}_{t} ))+\rho_{\textbf{f}_{t}}\Vert \textbf{f}_{t}(\textbf{v}_{t},\textbf{w}_{t} )\Vert_2^2\label{aug2}
\end{eqnarray}
\normalsize

Using (\ref{aug1})-(\ref{aug2}), we formulate Algorithm \ref{algo2} wherein both the above defined Lagrangian are alternately minimized on lines 2 and 4 respectively. From lines  5-8, the Lagrange multipliers are updated based on the residual of the projections and task constraints. On line 9, we increase the proximal weights by a positive factor, $\Delta >1$. The update stops when the optimization has converged or the iteration limit has been reached. The convergence is detected when residuals of projection and task constraints and change in magnitude of the cost function goes below the desired threshold.

Each minimization in algorithm \ref{algo2} is a convex QP problem with simple box bounds on the optimization variables and thus, can be solved analytically. To be precise, the process involves solving a unconstrained QP followed by appropriate clipping of the magnitudes of the variables.

\small
\begin{algorithm}
 \caption{Update Rule for the Proposed Optimizer: Semi Distributive}\label{algo2}
    \begin{algorithmic}[1]
    \small
     \While{$k\leq maxiter$}
           \For{$t= 0:t_f$}
           
           \small
          \begin{eqnarray}
(\textbf{v}_{t}^{k+1},\textbf{w}_{t}^{k+1}) = \arg\min L_2 (\textbf{q}^{k}_{t}, \lambda_{\textbf{v}_{t}}^k, \lambda_{\textbf{w}_{t}}^k, \rho_{\textbf{v}_{t}}^k, \rho_{\textbf{w}_{t}}^k  )\nonumber\\\vert \textbf{v}_t\vert \leq 1, \vert \textbf{w}_t\vert \leq 1 \label{first_opt}
\end{eqnarray}
\normalsize
\EndFor

\State 
\small
\begin{eqnarray}
\textbf{q}_{t}^{k+1}..\textbf{q}_{t_f}^{k+1} = \arg\min \sum_i L_1(\textbf{q}_{t}, \textbf{v}_{t}^{k+1}, \textbf{w}_{t}^{k+1}, \lambda_{\textbf{q}_{t}}^k, \rho_{\textbf{q}_{t}}^k ) \nonumber  \\
\text{such that}, \textbf{q}_t\in \mathcal{C}_{\textbf{q}}\label{second_opt}
\end{eqnarray} 
\normalsize

\small
\State $\boldsymbol{\lambda}_{\textbf{q}_{t}}^{k+1} = \boldsymbol{\lambda}_{\textbf{q}_{t}}^{k}+\rho_{\textbf{q}_{t}}^{k}(\textbf{A}\textbf{q}_{t}^{k+1}-\arctan2(\frac{\textbf{w}_{t}^{k+1}}{\textbf{v}_{t}^{k+1}}))$

\State $\boldsymbol{\lambda}_{\textbf{v}_{t}}^{k+1} = \boldsymbol{\lambda}_{\textbf{v}_{t}}^{k}+\rho_{\textbf{v}_{t}}^k(\textbf{v}_{t}^{k+1}- \textbf{b}_{v_{t}}(\textbf{q}_{t}^{k+1}))$  

\State   $\boldsymbol{\lambda}_{\textbf{w}_{t}}^{k+1} = \lambda_{\textbf{w}_{t}}^{k}+\rho_{\textbf{w}_{t}}^k(\textbf{w}_{t}^{k+1}- \textbf{b}_{w_{t}}(\textbf{q}_{t}^{k+1}))$         

\State $\boldsymbol{\lambda}_{\textbf{f}_{t}}^{k+1} = \boldsymbol{\lambda}_{\textbf{f}_{t}}^{k}+\rho_{\textbf{f}_{t}}^k(\textbf{f}_{t}(\textbf{v}_{t}^{k+1},\textbf{w}_{t}^{k+1}))$

\State $\rho_{\textbf{q}_{t}}^{k+1} = \rho_{\textbf{q}_{t}}^{k}\Delta$, $\rho_{\textbf{v}_{t}}^{k+1} = \rho_{\textbf{v}_{t}}^{k}\Delta$, $\rho_{\textbf{w}_{t}}^{k+1} = \rho_{\textbf{w}_{t}}^{k}\Delta$, $\rho_{\textbf{f}_{t}}^{k+1} = \rho_{\textbf{f}_{t}}^{k}\Delta$
\normalsize

\State If converged then break and exit.
           
\EndWhile 
\normalsize          
        \end{algorithmic}  
        \end{algorithm}
        
\normalsize

\noindent \textbf{Distributed Computation}: The structure of the optimization (\ref{first_opt}) in Algorithm \ref{algo2} is exactly same as that of (\ref{first_proj}) and thus can be parallelized across $n$ different processors. In contrast, the structure of optimization (\ref{second_opt}) and \ref{second_proj} are strikingly different. In (\ref{second_opt}), the cost function, $J(\textbf{q}_{t_i-2:t_i})$ introduces strong coupling between the variables and thus, it becomes imperative to formulate a large optimization where all the joint configurations at different time instants are computed simultaneously. 

To induce a stronger distributiveness, we propose a workaround by formulating a small approximation for the cost function. At the $k+1$ iteration, the cost function is simplified in the following manner. As can be seen, (\ref{sum_acc_approx}) gets rid of variable coupling and we show later that it is still rich enough to result in smooth joint motions.

\small
\begin{equation}
\widetilde{J}(\textbf{q}_{t-2:t}) = \Vert \textbf{q}_{t-2}^{k}+\textbf{q}_{t}-2\textbf{q}_{t-1}^k\Vert_2^2
\label{sum_acc_approx}.
\end{equation}
\normalsize

\noindent Using, \ref{sum_acc_approx}, we can replace  replace line 4 in the Algorithm \ref{algo2} with the optimization presented in Algorithm \ref{algo4}. Therein, $\widetilde{L}_1(.)$ represents the augmented Lagrangian (\ref{aug2}) with the approximate cost function (\ref{sum_acc_approx}).

\begin{algorithm}
 \caption{Replacement for line 4 of Algorithm \ref{algo2} to achieve full distributiveness}\label{algo4}
    \begin{algorithmic}[1]
    \For{$t= 0:t_f$}
    \small    
\begin{eqnarray}
\textbf{q}_{t}^{k+1}  = \widetilde{L}_1(\textbf{q}_{t}, \textbf{v}_{t}^{k+1}, \textbf{w}_{t}^{k+1}, \boldsymbol{\lambda}_{\textbf{q}_{t}}^k, \rho_{\textbf{q}_{t}}^k ) \nonumber  \\
\text{such that}, \textbf{q}_t\in \mathcal{C}_{\textbf{q}}\label{second_opt_2}
\end{eqnarray}
\normalsize
\EndFor
    \end{algorithmic}  
        \end{algorithm}

\subsection{Proposed Optimizer with Task Constraints, (\ref{nonlin_gen_spatial})}
\noindent We begin by pointing out that to extend Algorithm \ref{algo2} to handle  task constraints, (\ref{nonlin_gen_spatial}), just the lines 2-3 of Algorithm \ref{algo2}, in particular the optimization (\ref{first_opt}) needs to be replaced with something more general. The rest of the Algorithm remains exactly the same. To this end, we first present the following reformulation of (\ref{nonlin_gen_spatial}):

${^d}f_t(v_t^j, w_t^j, \overline{\textbf{v}}_t^j, \overline{\textbf{w}}_t^j)=$
\small
\begin{equation}
  \begin{cases}
    {^d}g_t^{1}(\overline{\textbf{v}}_t^1, \overline{\textbf{w}}_t^1)v_t^1+{^d}h_t^{1}(\overline{\textbf{v}}_t^1, \overline{\textbf{w}}_t^1)w_t^1+{^d}p_t^1(\overline{\textbf{v}}_t^1, \overline{\textbf{w}}_t^1)\\
    {^d}g_t^2(\overline{\textbf{v}}_t^2, \overline{\textbf{w}}_t^2)v_t^2+{^d}h_t^2(\overline{\textbf{v}}_t^2, \overline{\textbf{w}}_t^2)w_t^2+{^d}p_t^2(\overline{\textbf{v}}_t^2, \overline{\textbf{w}}_t^2)\\
...................................\\
{^d}g_t^m(\overline{\textbf{v}}_t^m, \overline{\textbf{w}}_t^m)v_t^m+{^d}h_t^m(\overline{\textbf{v}}_t^m, \overline{\textbf{w}}_t^2)w_t^m+{^d}p_t^m(\overline{\textbf{v}}_t^m, \overline{\textbf{w}}_t^m)\\    
  \end{cases}
  \label{nonlin_gen_spatial2}
\end{equation}
\normalsize

\noindent Where,  we have used the transformation
\small
\begin{equation}
v_t^j = \cos(q_t^j), w_t^j = \sin(q_t^j)
\label{transform3} 
\end{equation}
\normalsize

\noindent In (\ref{nonlin_gen_spatial2}), $\overline{\textbf{v}}_t^j, \overline{\textbf{w}}_t^j$ are obtained from $\textbf{v}_t, \textbf{w}_t$ by removing the $j^{th}$ element. Furthermore, special attention should be paid to the arguments of functions, ${^d}g_t^j(.)$, ${^d}h_t^j(.)$ and ${^d}p_t^j(.)$. As mentioned earlier (recall discussions around (\ref{nonlin_gen_spatial}) ), these functions depend on sine and cosine of all the joint angles except $q_t^j$. So, in  (\ref{nonlin_gen_spatial2}), special care has been taken to explicitly highlight the fact that all those sine and cosine have now been replaced with the help of (\ref{transform3}).

It is clear that (\ref{nonlin_gen_spatial2}) is  affine with respect to each separate pairs of ($v_t^j, w_t^j$)   but is highly non-linear and non-convex when all the pairs are considered jointly. This is the classic structure prevalent in many non-convex optimization problems which are shown to be efficiently solvable with AM techniques (recall Section \ref{prelim}) \cite{boyd_scp}. We build on this insight and propose Algorithm \ref{algo5} which provides a replacement for lines 2-3 in Algorithm \ref{algo2}. The augmented Lagrangian, $L_3$ is a simpler variant of that presented in (\ref{aug2_general}) and herein, the multipliers, $\lambda_{v_{t}^j}$, $\lambda_{w_{t}^j}$ are scalars. In fact, they are the $j^{th}$ component of $\boldsymbol{\lambda}_{\textbf{v}_t}$ and $\boldsymbol{\lambda}_{\textbf{w}_t}$ respectively. $\lambda_{\textbf{f}_t}$ has the same definition as in (\ref{aug2}). Similarly, the proximal weights, $\rho_{\textbf{v}_t}$  and $\rho_{\textbf{w}_t}$ can be directly inherited from (\ref{aug2}).

\small
\begin{eqnarray}
L_3({v}_{t}^j, {w}_{t}^j ) = \lambda_{v_{t}^j}(v_t^j- \cos(q_t^j)+\rho_{\textbf{v}_t}(v_t^j- \cos(q_t^j)^2\nonumber \\
+\lambda_{w_{t}^j}(w_t^j- \sin(q_t^j)+\rho_{\textbf{w}_t}(w_t^j- \sin(q_t^j)^2\nonumber\\
+\boldsymbol{\lambda}_{\textbf{f}_{t}}^T(\textbf{f}_{t}(v_j^t, w_j^t,\overline{\textbf{v}}_{t},\overline{\textbf{w}}_{t} ))+\rho_{\textbf{f}_{t}}\Vert \textbf{f}_{t}(v_j^t, w_j^t, \overline{\textbf{v}}_{t},\overline{\textbf{w}}_{t} )\Vert_2^2\label{aug2_general}
\end{eqnarray}
\normalsize

\begin{algorithm}
 \caption{Replacement for line 2-3 of Algorithm \ref{algo2} to handle general task constraints}\label{algo5}
    \begin{algorithmic}[1]
    \For{$t= 0:t_f$}
    \For{$j= 0:m$}    
    \State ${^d}g_t^j \leftarrow {^d} g_t^j((\overline{\textbf{v}}_t^j)^k,(\overline{\textbf{w}}_t^j)^k)$, ${^d}h_t^j \leftarrow {^d} h_t^j((\overline{\textbf{v}}_t^j)^k,(\overline{\textbf{w}}_t^j)^k)$, ${^d}p_t^j \leftarrow {^d} p_t^j((\overline{\textbf{v}}_t^j)^k,(\overline{\textbf{w}}_t^j)^k)$
 \small
\begin{eqnarray}\nonumber
(v_{t}^j)^{k+1},(w_{t}^j)^{k+1} = \arg\min L_3 ( (q_t^j)^k, \lambda_{{v}_{t}^j}^k, \lambda_{{w}_{t}^j}^k, \rho_{{v}_{t}^j}^k, \rho_{{w}_{t}^j}^k  )\label{first_opt_general}
\end{eqnarray}
\normalsize
\State Use $(v_t^j)^{k+1},(w_t^j)^{k+1}$ to compute $(\overline{\textbf{v}}_t^{j+1})^{k+1}$, $(\overline{\textbf{w}}_t^{j+1})^{k+1}$
\EndFor
\EndFor
\State Stack $(v_t^j)^{k+1}, (w_t^j)^{k+1}$ to form $(\textbf{v}_t)^{k+1}, (\textbf{w}_t)^{k+1}$
    \end{algorithmic}  
        \end{algorithm}

%
%
%
%
%
%
%

\section{Applications}


\subsection{Planar Manipulator}

\subsubsection{Application 1} As our first application, we consider the task shown in Fig.\ref{planar_circle_config} where a 6 $dof$ planar manipulator is required to execute a circular trajectory while always keeping the orientation of the last link parallel to the horizontal. The task constraints for this application are given by the following non-linear equalities. As can be seen, (\ref{planar_task1})-(\ref{planar_task3}) is in the form given by (\ref{nonlin_gen}).

\small
\begin{subequations}
\begin{align}
a_1\cos(q_t^1)+a_2\cos(q_t^1+q_t^2)+..a_6\cos(q_t^1+..q_t^6)-x_t^{des} = 0 \label{planar_task1} \\
a_1\sin(q_t^1)+a_2\sin(q_t^1+q_t^2)+..a_6\sin(q_t^1+..q_t^6)-y_t^{des} = 0\label{planar_task2}\\
\cos(q_t^1+..q_t^6) = 1, \sin(q_t^1+..q_t^6) = 0 \label{planar_task3}
\end{align}
\end{subequations}
\normalsize

\noindent The results are summarized in Fig.\ref{planar_circle_config}-(\ref{planar_circle_thetadot}). Following important aspects should be noted. Firstly, Fig. \ref{planar_circle_proj} shows that the projection residual, i.e $\Vert\textbf{v}_t-\textbf{b}_{\textbf{v}_t(\textbf{q}_t)}\Vert_1$ and $\Vert\textbf{w}_t-\textbf{b}_{\textbf{w}_t(\textbf{q}_t)}\Vert_1$ at all time instants goes to zero as the optimization progresses. Secondly, the joint angles and their velocities shown in Fig.\ref{planar_circle_joint} and \ref{planar_circle_thetadot} respectively, reveal that $\Vert\textbf{q}_0-\textbf{q}_n\Vert_1 = 0.001 rad$ and $\Vert\dot{\textbf{q}}_0-\dot{\textbf{q}}_n\Vert_1 = 0.01 rad/s$.  Thus, it can be seen that our proposed optimizer results in a closed cyclic trajectory in the joint space corresponding to a similar natured trajectory in the task space. It should be noted that this important result manifests automatically in our formulation without the need to incorporate any additional constraints aimed at this specific purpose.

%

\subsubsection{Application 2} 
\begin{figure*}[!h]
  \centering
   \subfigure[]{
    \includegraphics[width= 4.35cm, height=3.6cm] {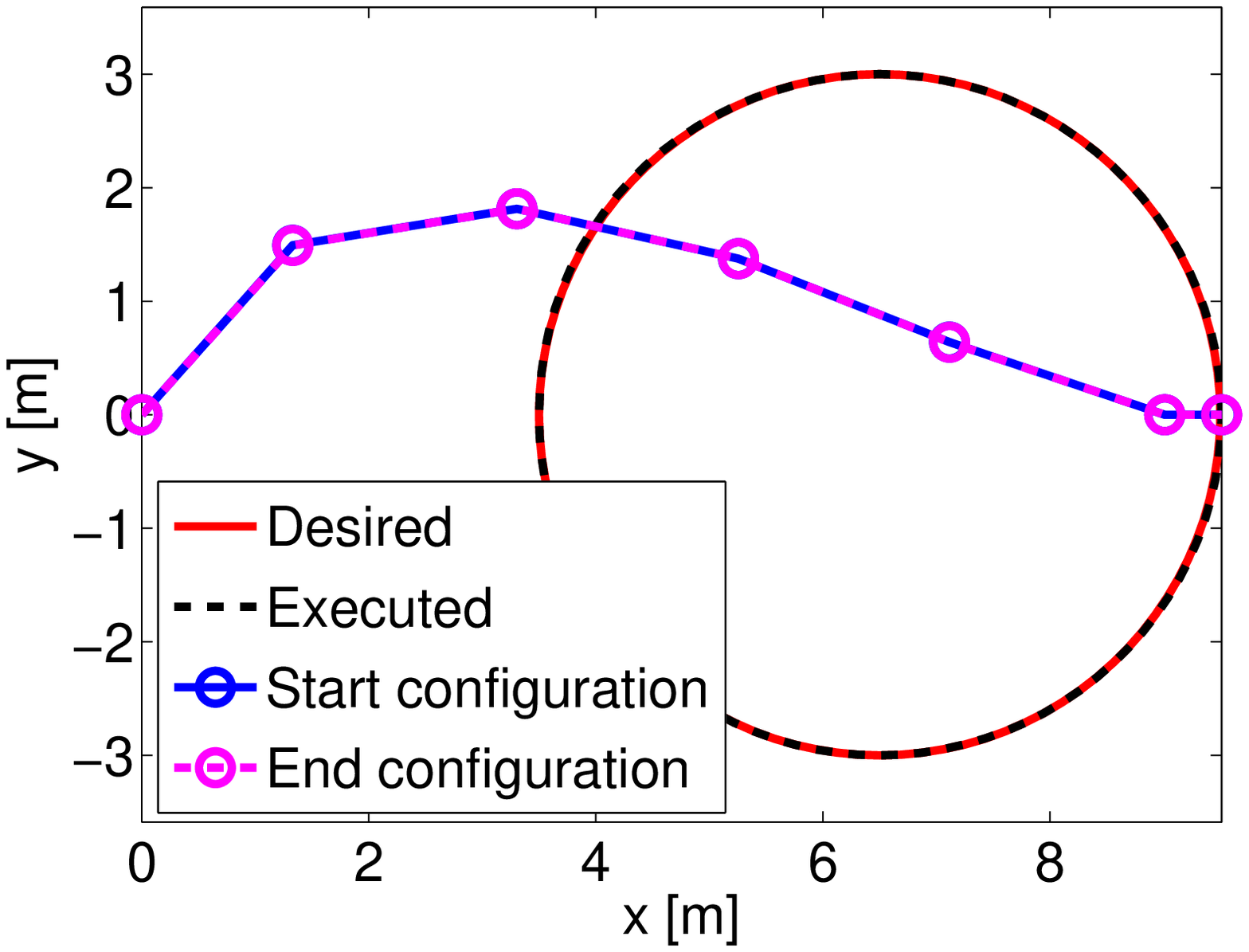}
    \label{planar_circle_config}
   }\hspace{-0.7cm}
   \subfigure[]{
    \includegraphics[width= 4.35cm, height=3.6cm] {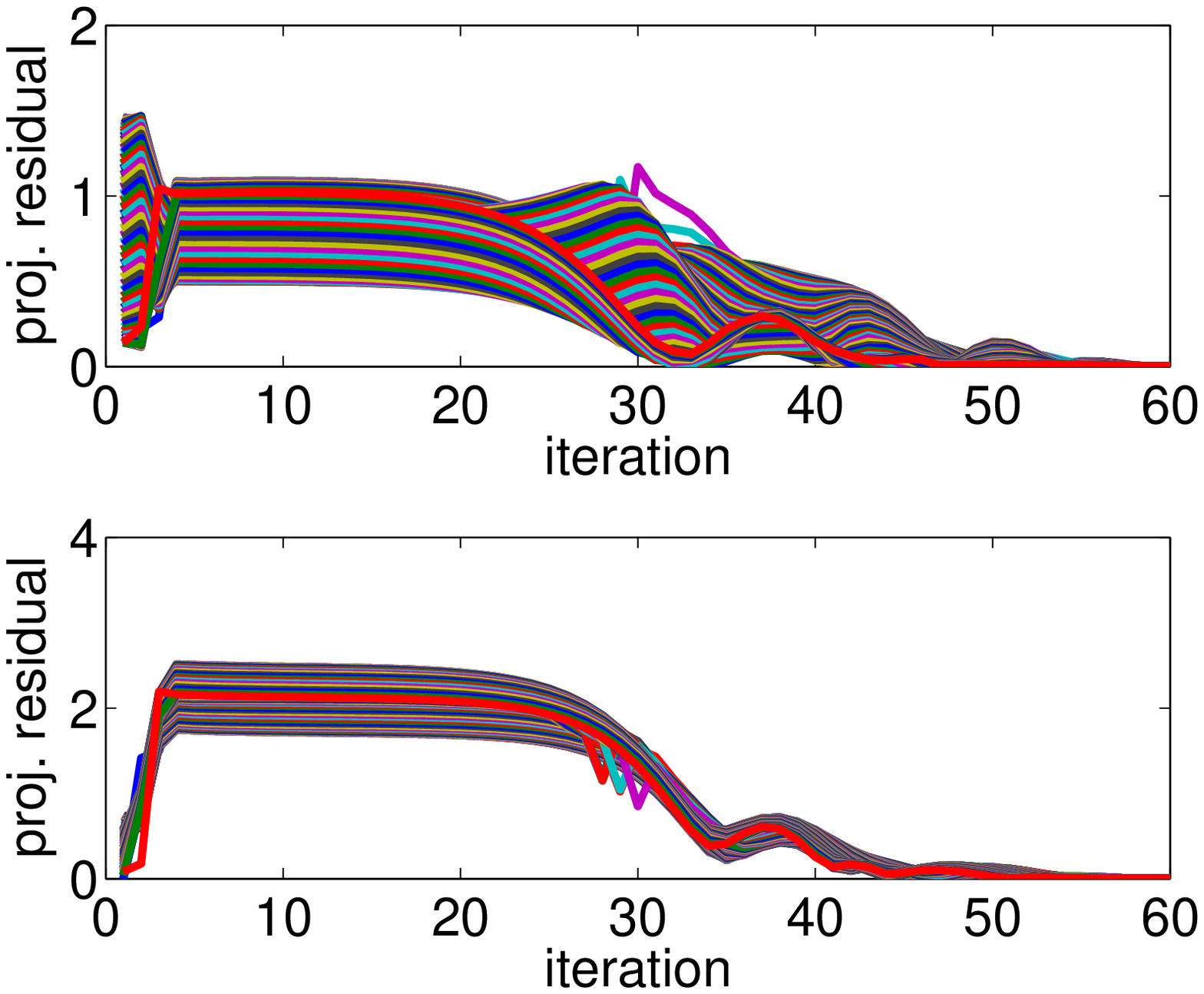}
    \label{planar_circle_proj}
   }\hspace{-0.7cm}
   \subfigure[]{
      \includegraphics[width= 4.35cm, height=3.6cm] {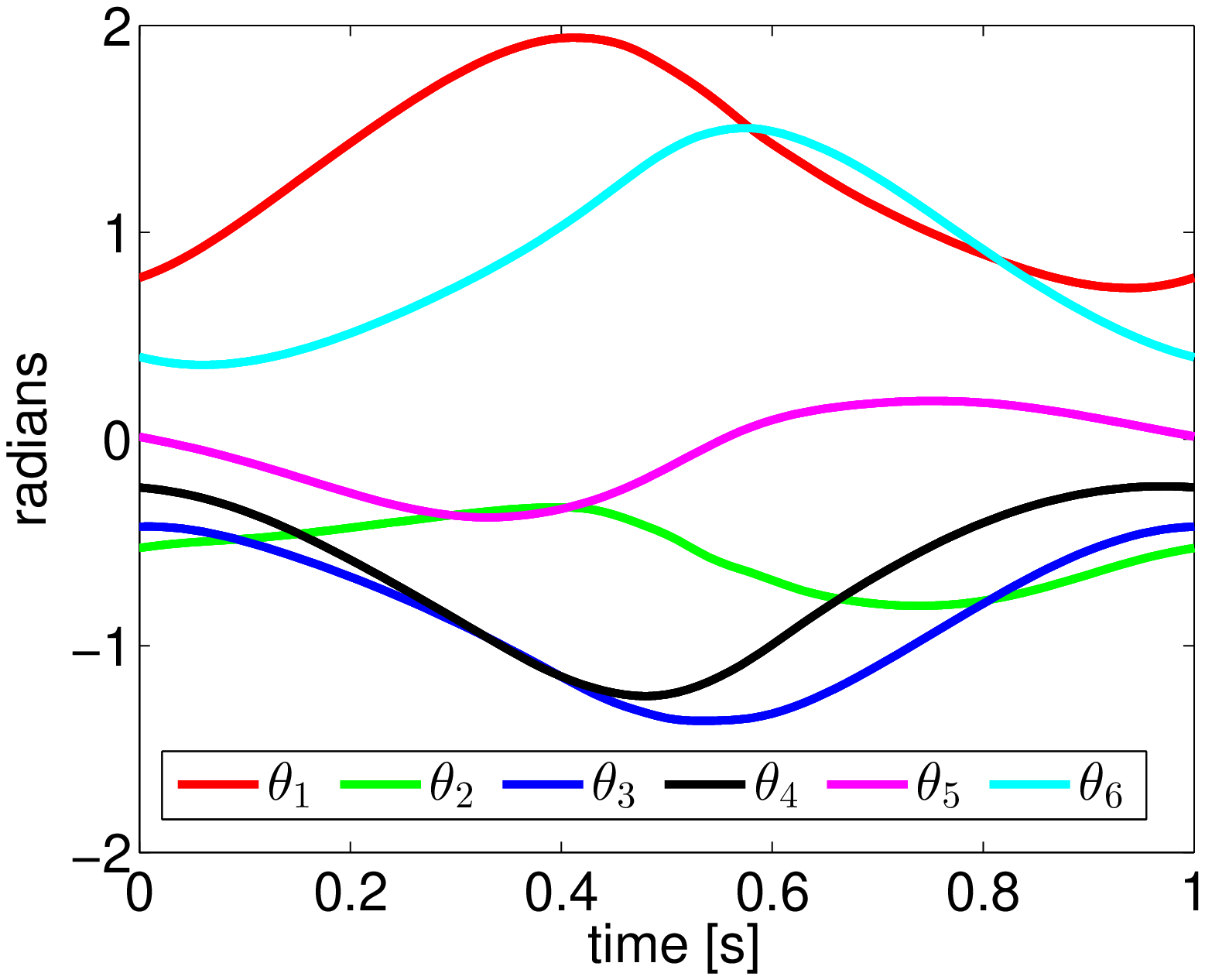}
    \label{planar_circle_joint}
   }\hspace{-0.5cm}
   \subfigure[]{
      \includegraphics[width= 4.35cm, height=3.6cm] {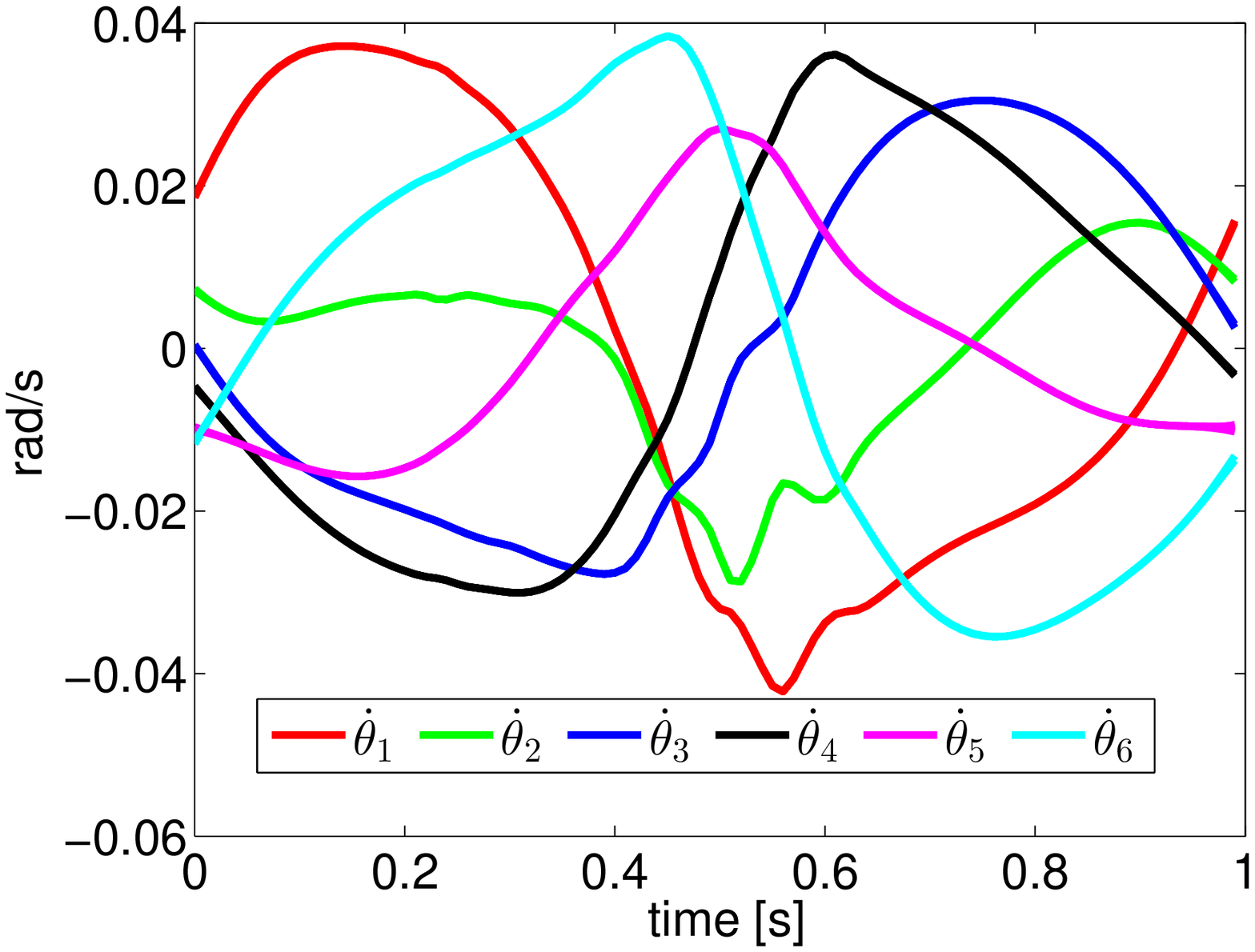}
    \label{planar_circle_thetadot}
   }
   \subfigure[]{
    \includegraphics[width= 4.35cm, height=3.6cm] {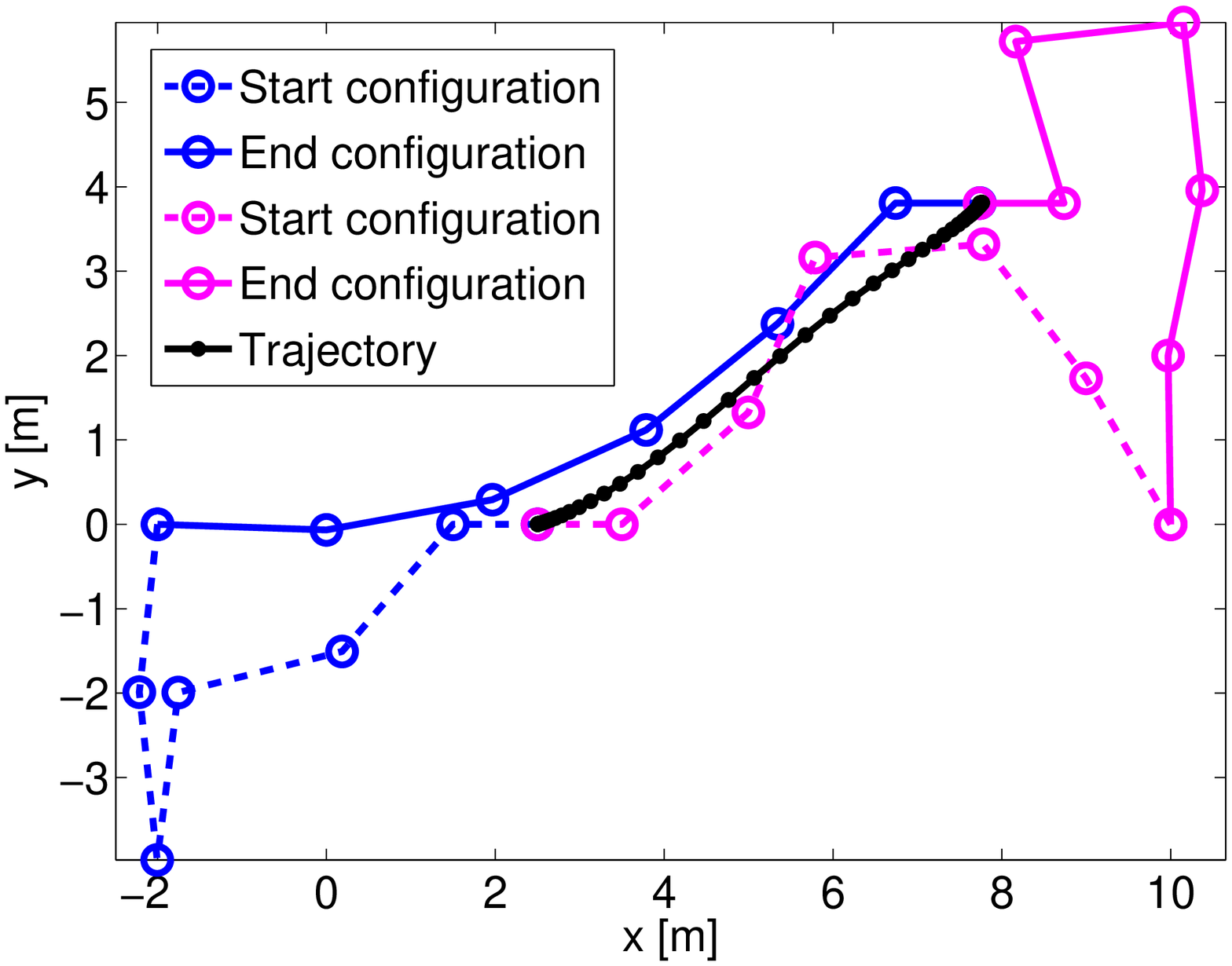}
    \label{planar_dual_config}
   }\hspace{-0.7cm}
   \subfigure[]{
    \includegraphics[width= 4.35cm, height=3.6cm] {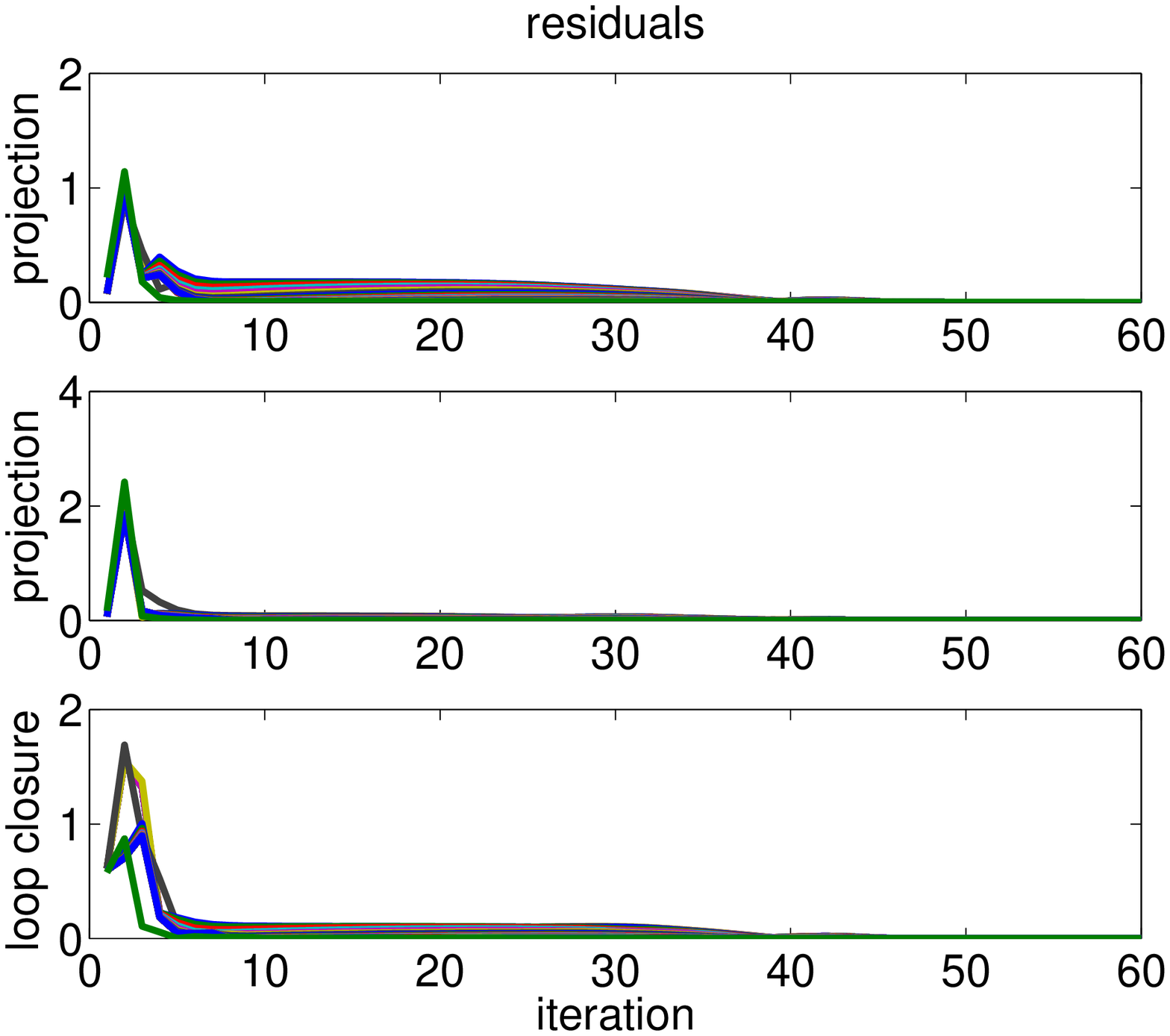}
    \label{planar_dual_proj}
   }
   \subfigure[]{
    \includegraphics[width= 4.35cm, height=3.6cm] {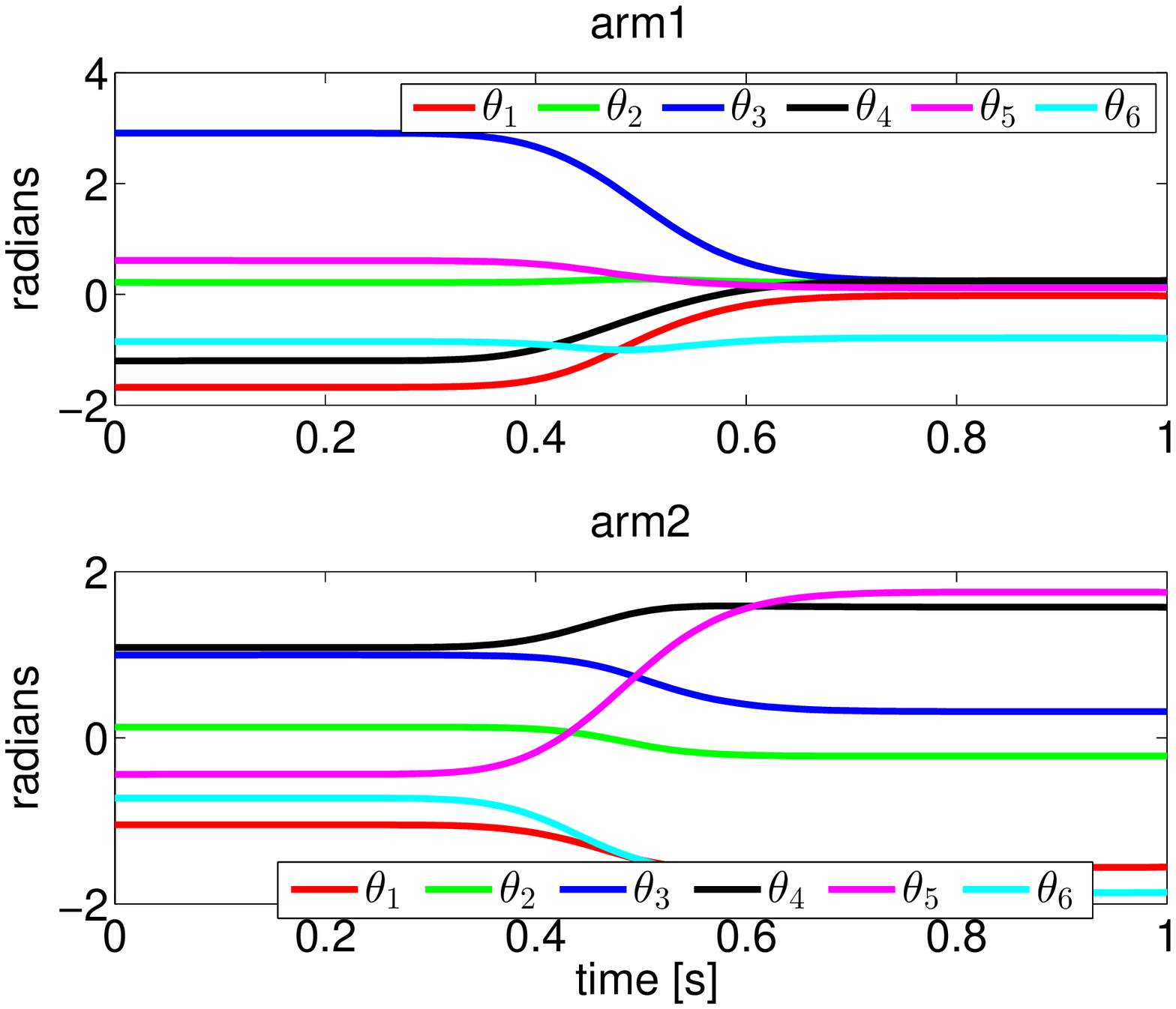}
    \label{planar_dual_joint}
   }\hspace{-0.7cm}
   \subfigure[]{
    \includegraphics[width= 4.35cm, height=3.6cm] {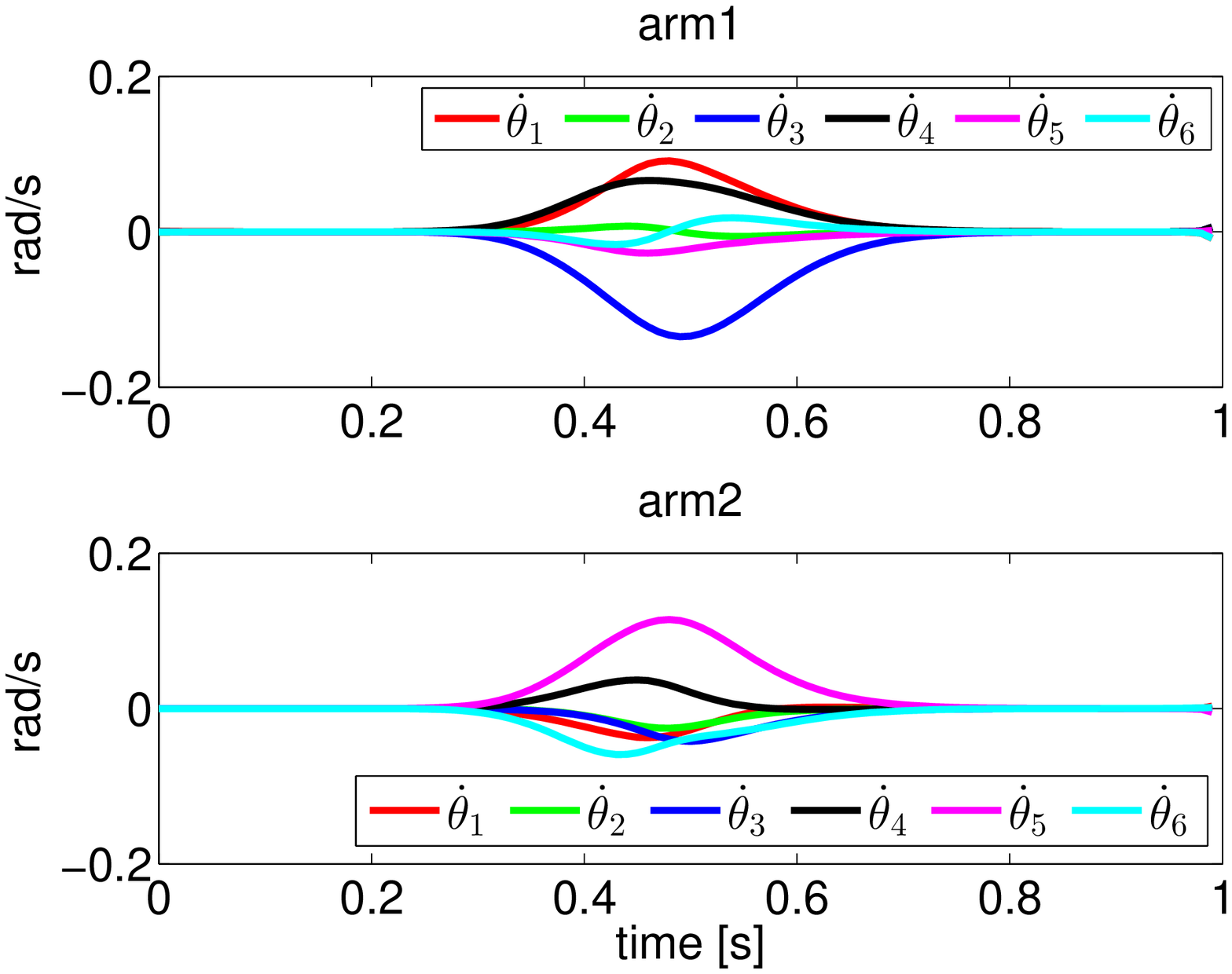}
    \label{planar_dual_jointvel}
   } 
  \caption{(a) Planar redundant manipulator executing a cyclic task space trajectory. (b) residuals of constraints, (\ref{slack1})-(\ref{slack2}). (c)-(d) plot of joint angles and velocities. (e) a smooth task space trajectory for redundantly actuated closed kinematic chain.  (f) top and middle plots shows the residual of (\ref{slack1})-(\ref{slack2}) with iteration while the bottom plot shows the residual of loop closure constraints for the closed kinematic chain. (g)-(h) joint angles and  velocities for the closed kinematic chain.  }   
  \vspace{-0.5cm}       
\end{figure*}

Our second application set up is shown in Fig.\ref{planar_dual_config} and concerns with simultaneously computing a joint and task space trajectory for point to point motion of a redundantly actuated closed kinematic chain formed by coupling two planar 6 $dof$ manipulator. The coupling is achieved by fixing the orientation of the last link of both the manipulators and joining them together. The task constraints for this application consists of the following parts: (i) loop closure constraints which manifests in the form of equality constraints ensuring that the end effector of both the manipulators are at the same place and parallel to the horizontal at all time instants. (ii) An equality constraint which ensures that the end effector of the first manipulator reaches the final position at the final time instant.

Due to space constraints, we do not present the mathematical form of these constraints. The results are summarized in Fig.\ref{planar_dual_config}-\ref{planar_dual_jointvel}. Fig.\ref{planar_dual_proj} shows the residual of projection (top and middle sub-figure) and loop closure constraints (bottom sub-figure). The joint angle and velocity profiles shown in Fig.\ref{planar_dual_joint}-\ref{planar_dual_jointvel} highlight the smoothness in the joint trajectory.

\subsubsection{Application 3} In the application shown in Fig.\ref{kuka_circle_config}, a KUKA LWR manipulator \cite{kuka} is required to execute a circular trajectory in the $Y-Z$ plane while keeping the orientation fixed at $\textbf{R}_{XYZ}(0,0,0)$. The task constraints has the following components (i) Three equality constraints for the position of the end-effector. (ii) Nine equality constraints corresponding to the nine elements of the rotation matrix. The projection residuals are shown in Fig.\ref{kuka_circle_proj}. Since, the task space trajectory is cyclic, the joint angles and their velocities shown in Fig.\ref{kuka_circle_joint} reveal that $\Vert\textbf{q}_0-\textbf{q}_n\Vert_1 = 0.00096 rad$ and $\Vert\dot{\textbf{q}}_0-\dot{\textbf{q}}_n\Vert_1 = 0.06 rad/s$. Thus, again, we obtain a cyclic trajectory in the joint space for a task space trajectory of the same nature. Fig.\ref{kuka_circle_posres} shows the residual of the task constraints.

\subsubsection{Application 4} In this application, a KUKA LWR manipulator is required to move between a given start and goal position on the $Y-Z$ plane while maintaining a fixed orientation of $\textbf{R}_{XYZ}( 0.001,    1.04,  0.007 ) rad.$ The task constraint components include (i) An equality constraint constraining the motion on the $Y-Z$ plane. (ii) Three equality constraints for the final position and (iii) Nine equality constraints for the nine elements of the rotation matrix.

Fig.\ref{kuka_path2} show the computed trajectory. Fig.\ref{kuka_path2_proj} shows the projection residuals. Fig.\ref{kuka_path1_joint} shows the joint angles and their velocities. Fig.\ref{kuka_path2_posres} shows the residual of the position and orientation constraints on the end-effector.

\subsubsection{Application 5} 
\begin{figure*}[!h]
  \centering
 \subfigure[]{
    \includegraphics[width= 4.35cm, height=3.6cm] {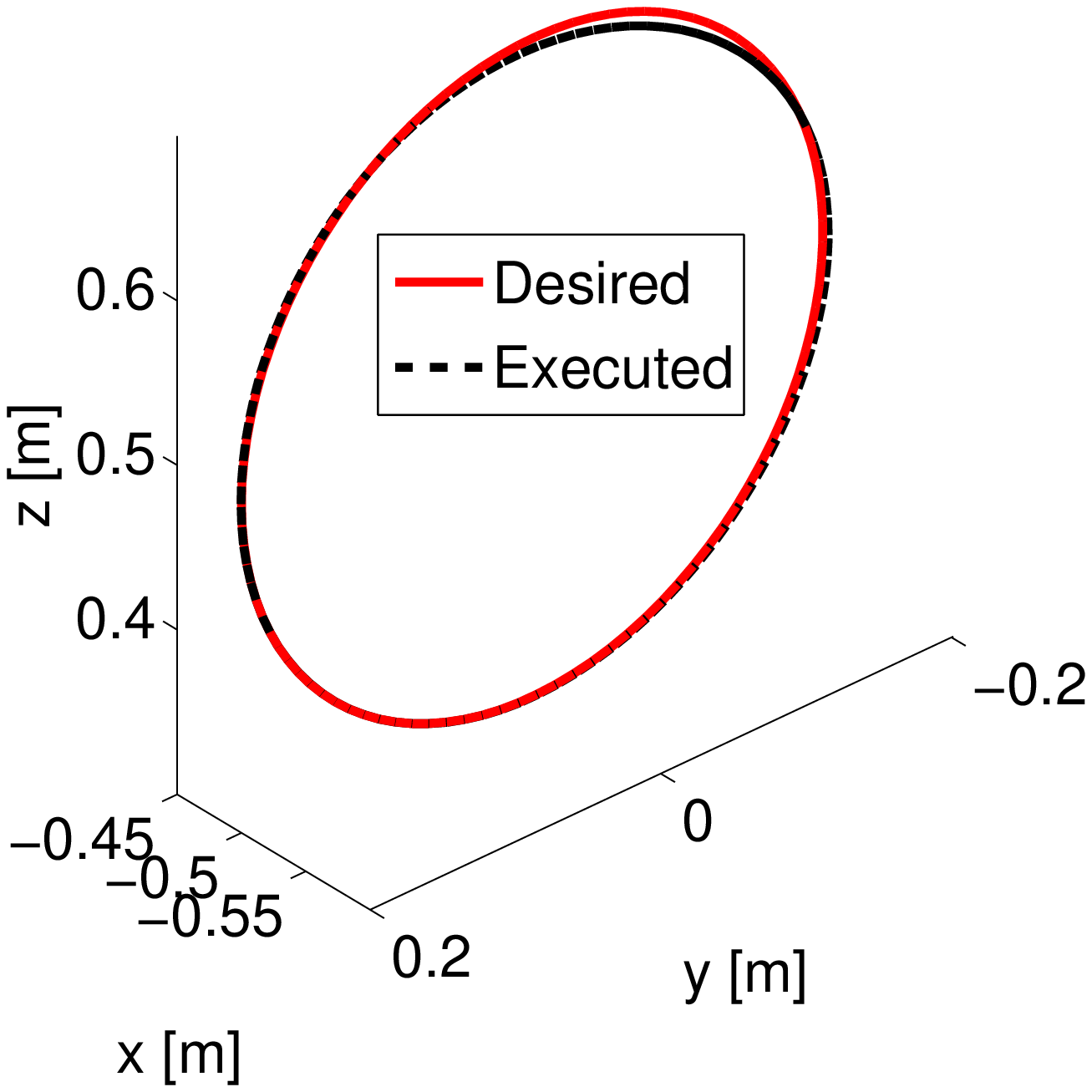}
    \label{kuka_circle_config}
   }\hspace{-0.7cm}
   \subfigure[]{
    \includegraphics[width= 4.35cm, height=3.6cm] {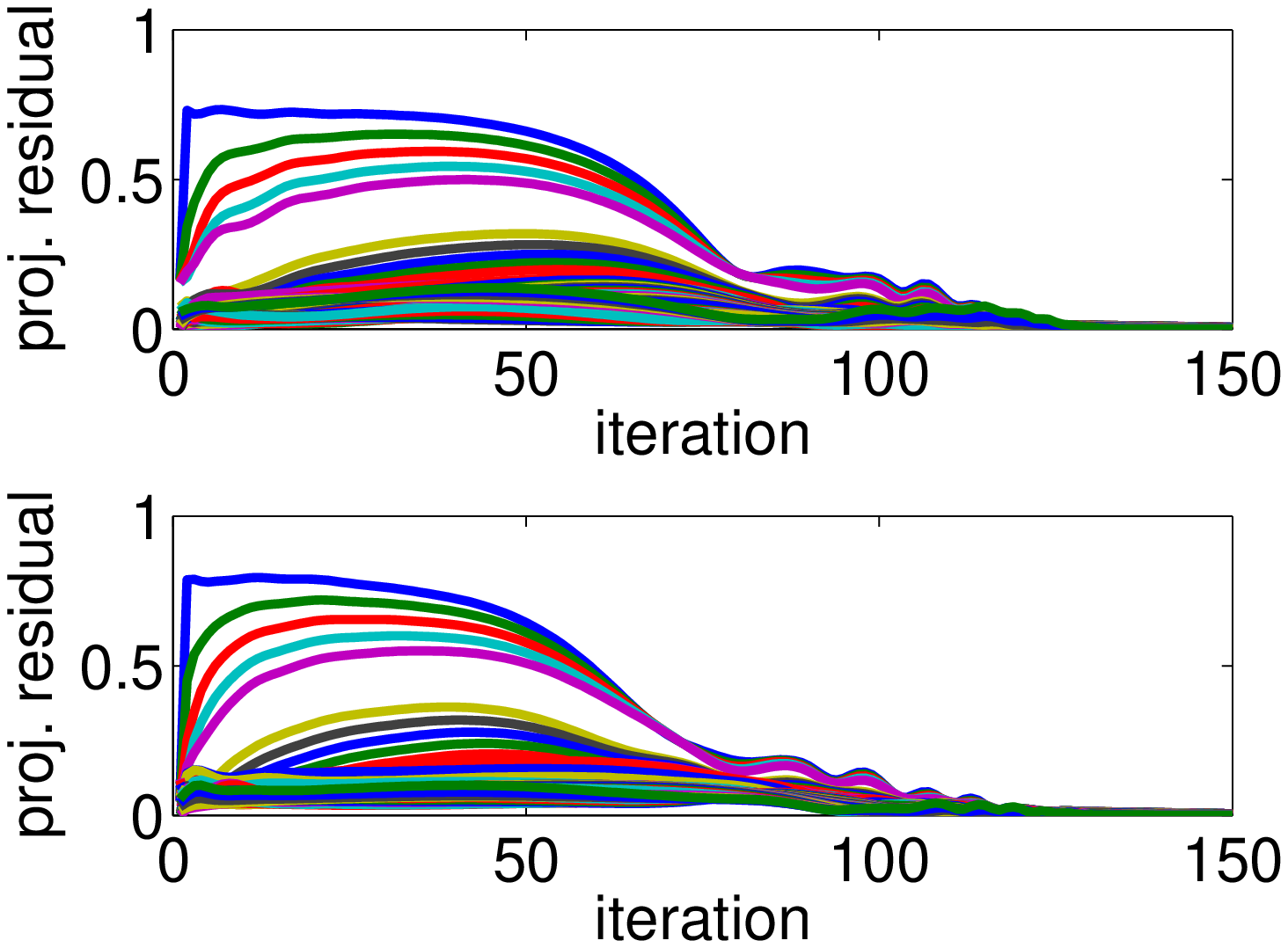}
    \label{kuka_circle_proj}
   }\hspace{-0.7cm}
   \subfigure[]{
      \includegraphics[width= 4.35cm, height=3.6cm] {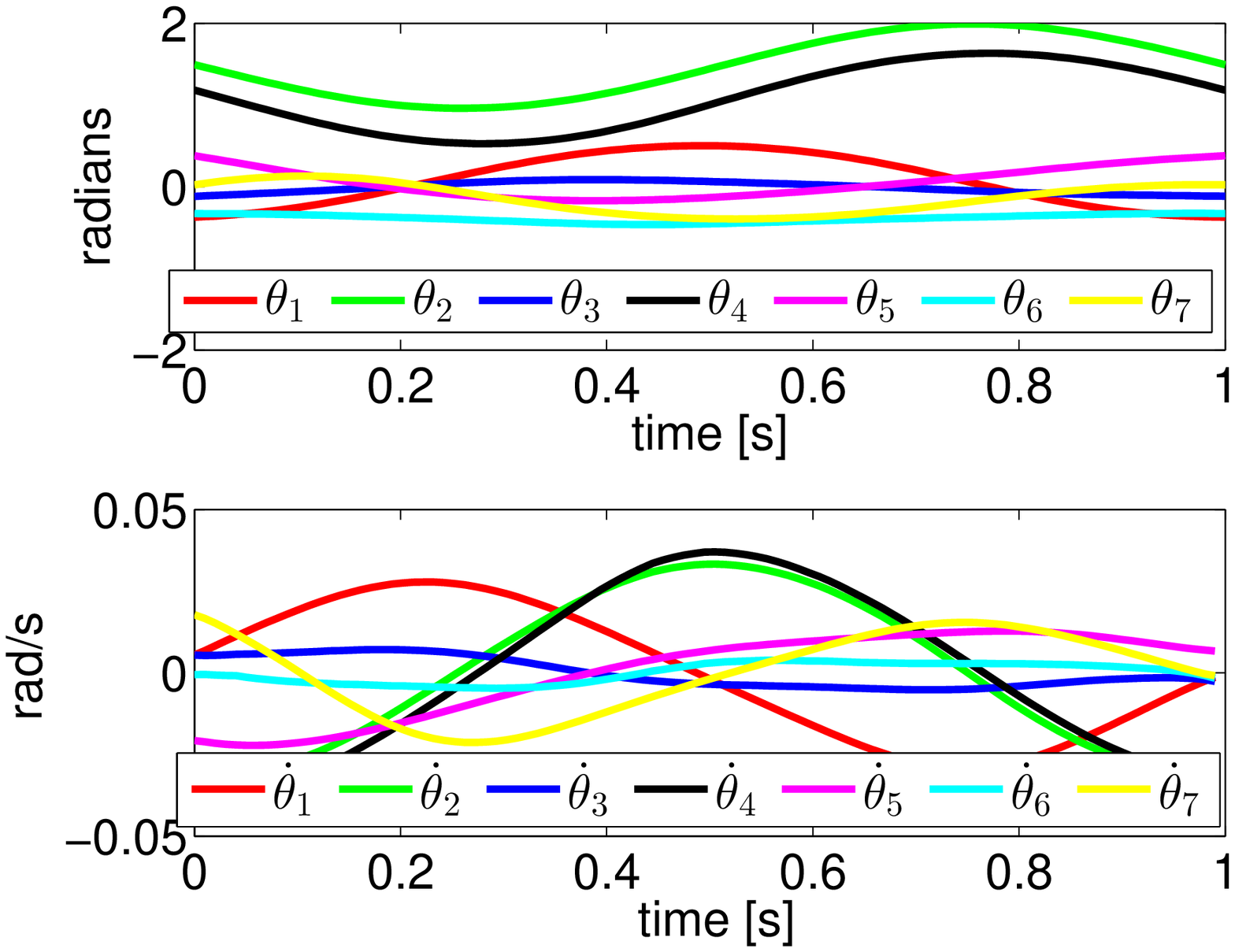}
    \label{kuka_circle_joint}
   }\hspace{-0.5cm}
   \subfigure[]{
      \includegraphics[width= 4.35cm, height=3.6cm] {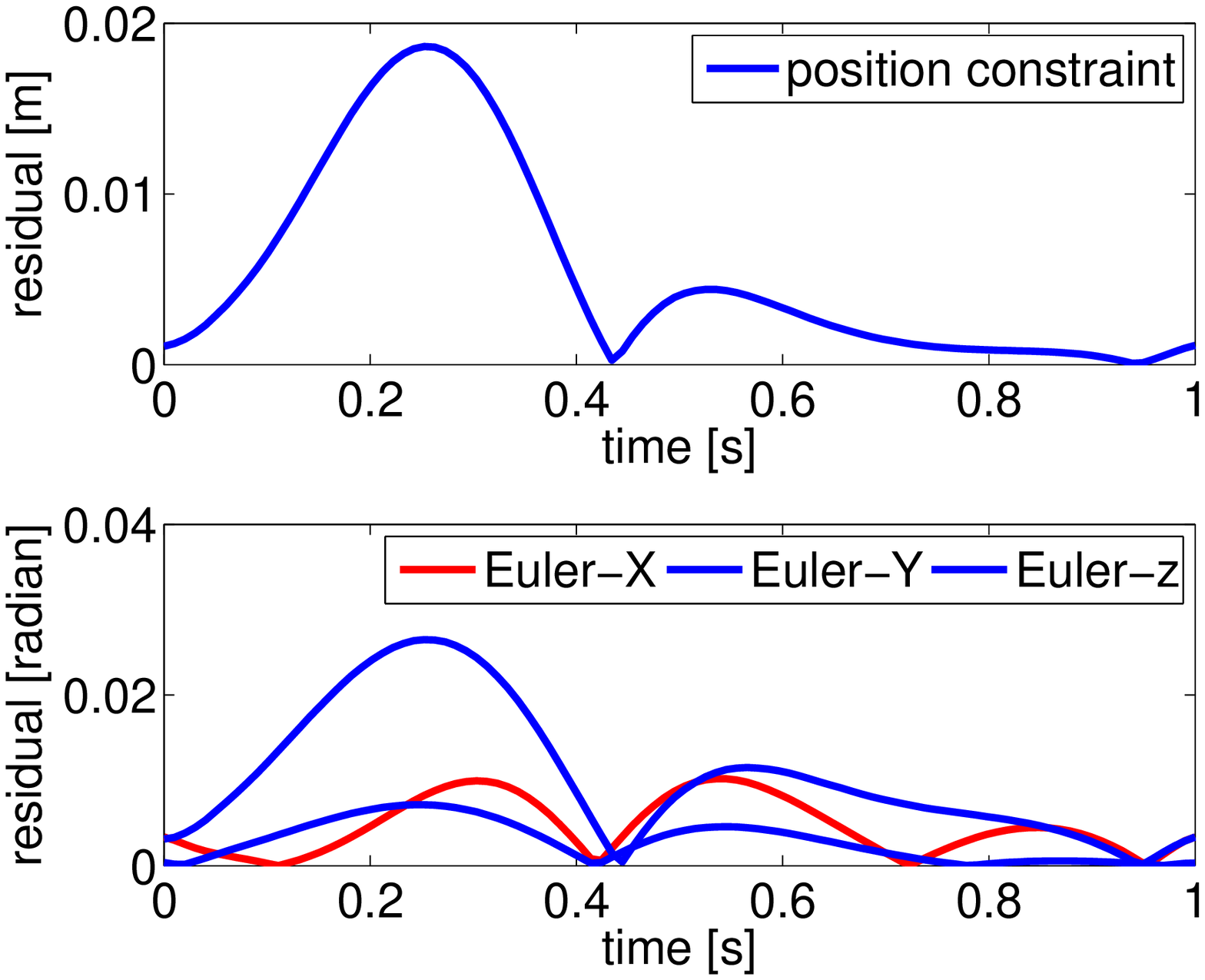}
    \label{kuka_circle_posres}
   }\hspace{-0.5cm}
   \subfigure[]{
    \includegraphics[width= 4.35cm, height=3.6cm] {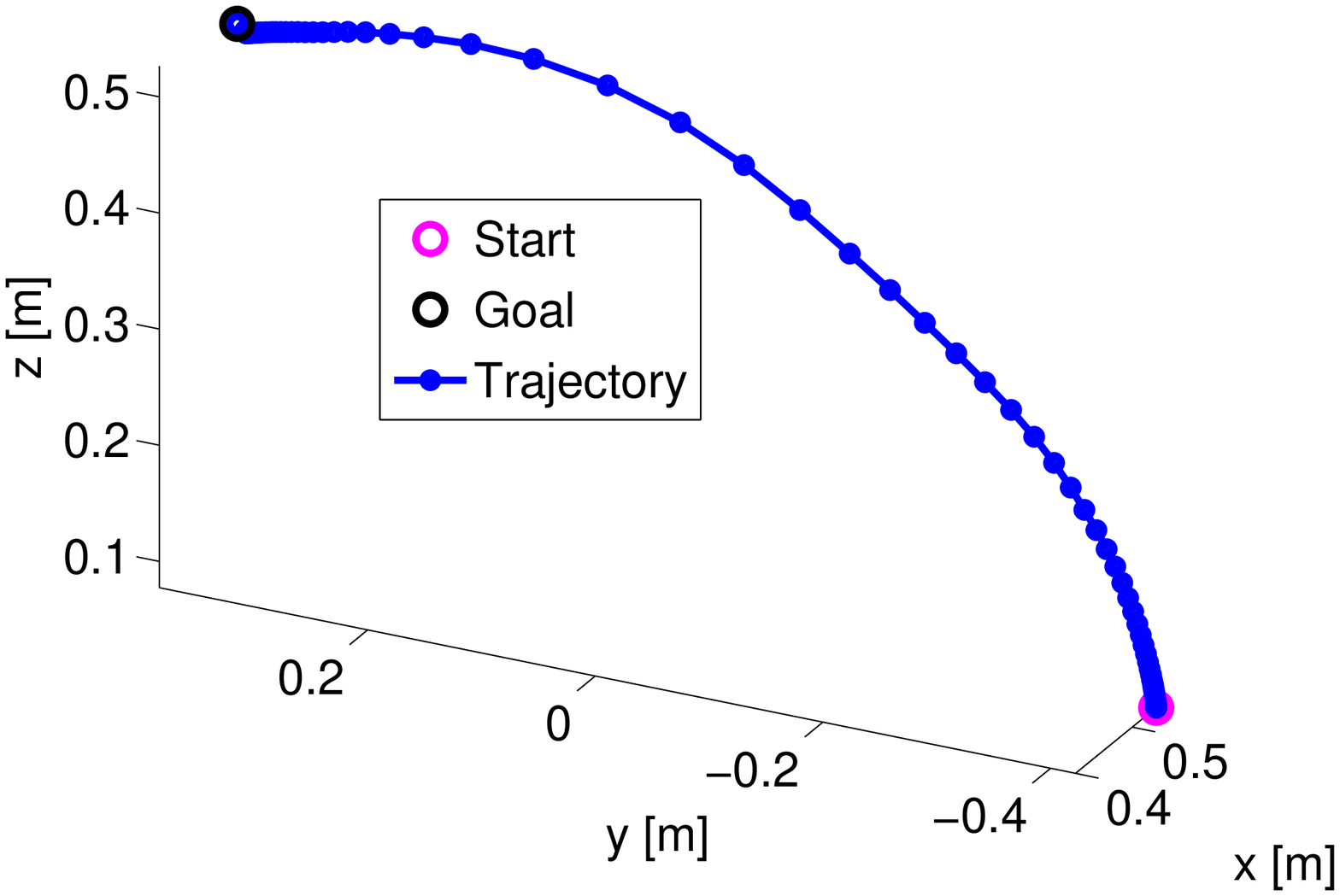}
    \label{kuka_path2}
   }\hspace{-0.7cm}
   \subfigure[]{
    \includegraphics[width= 4.35cm, height=3.6cm] {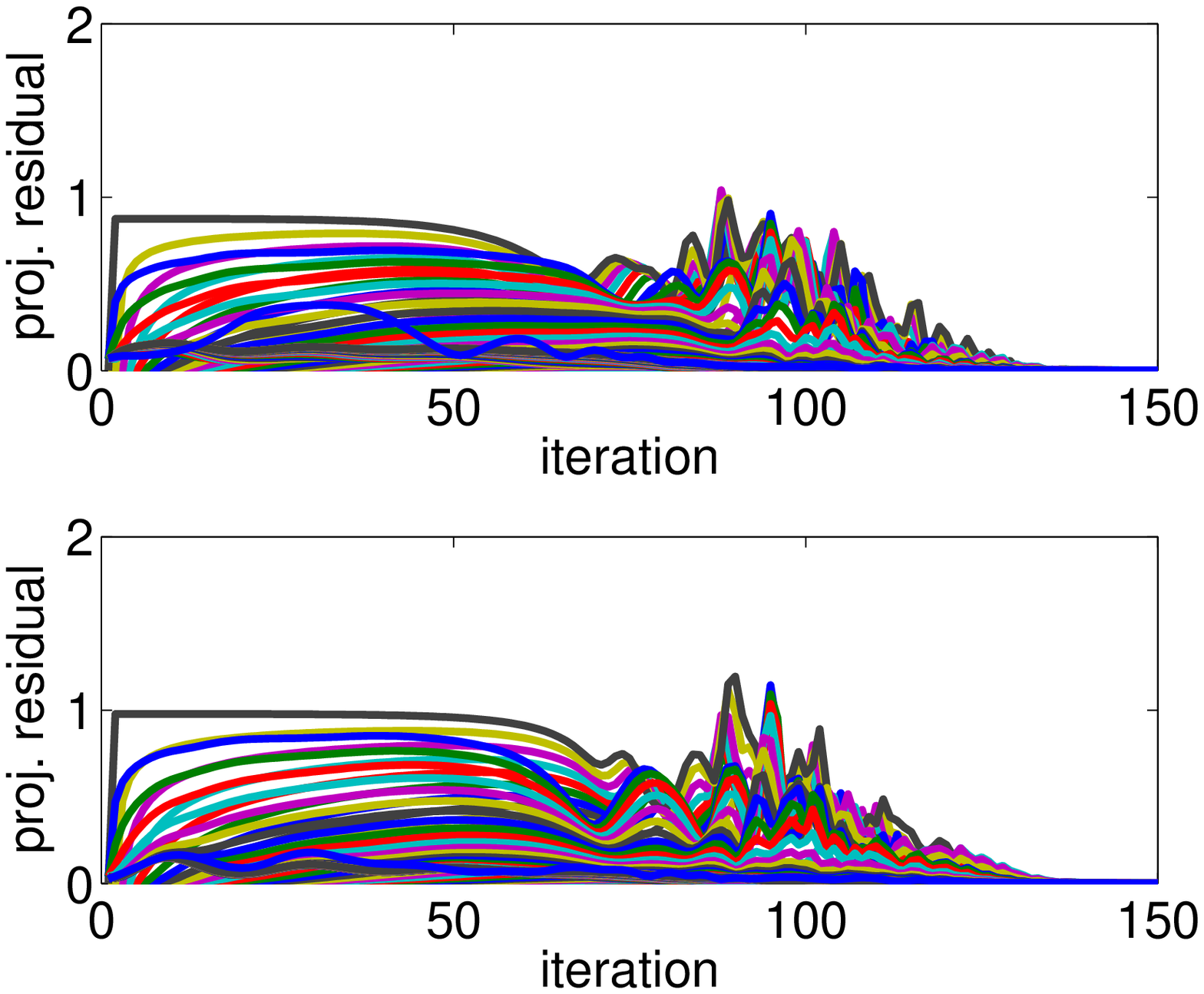}
    \label{kuka_path2_proj}
   }\hspace{-0.7cm}
   \subfigure[]{
      \includegraphics[width= 4.35cm, height=3.6cm] {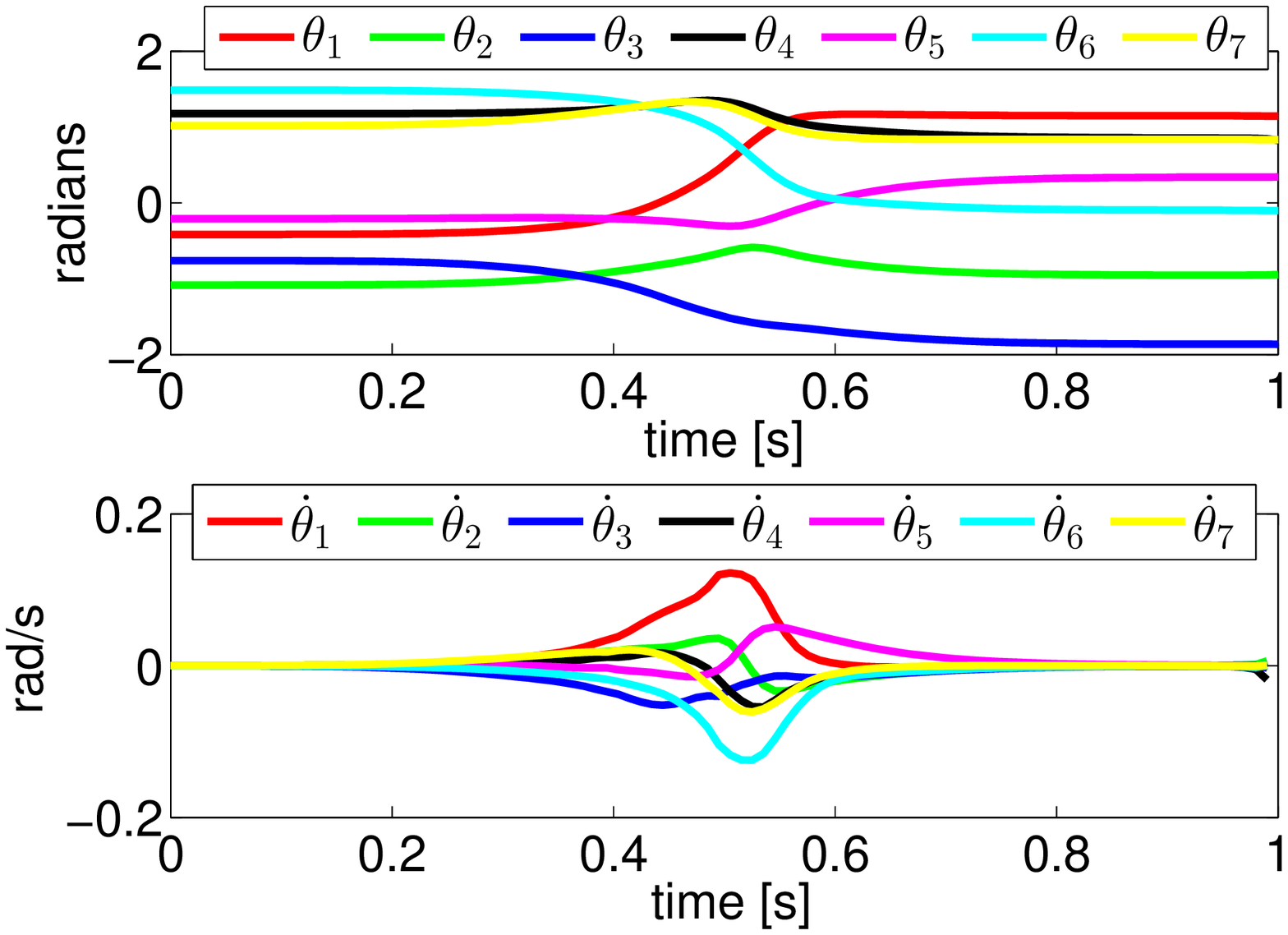}
    \label{kuka_path2_joint}
   }\hspace{-0.5cm}
   \subfigure[]{
      \includegraphics[width= 4.35cm, height=3.6cm] {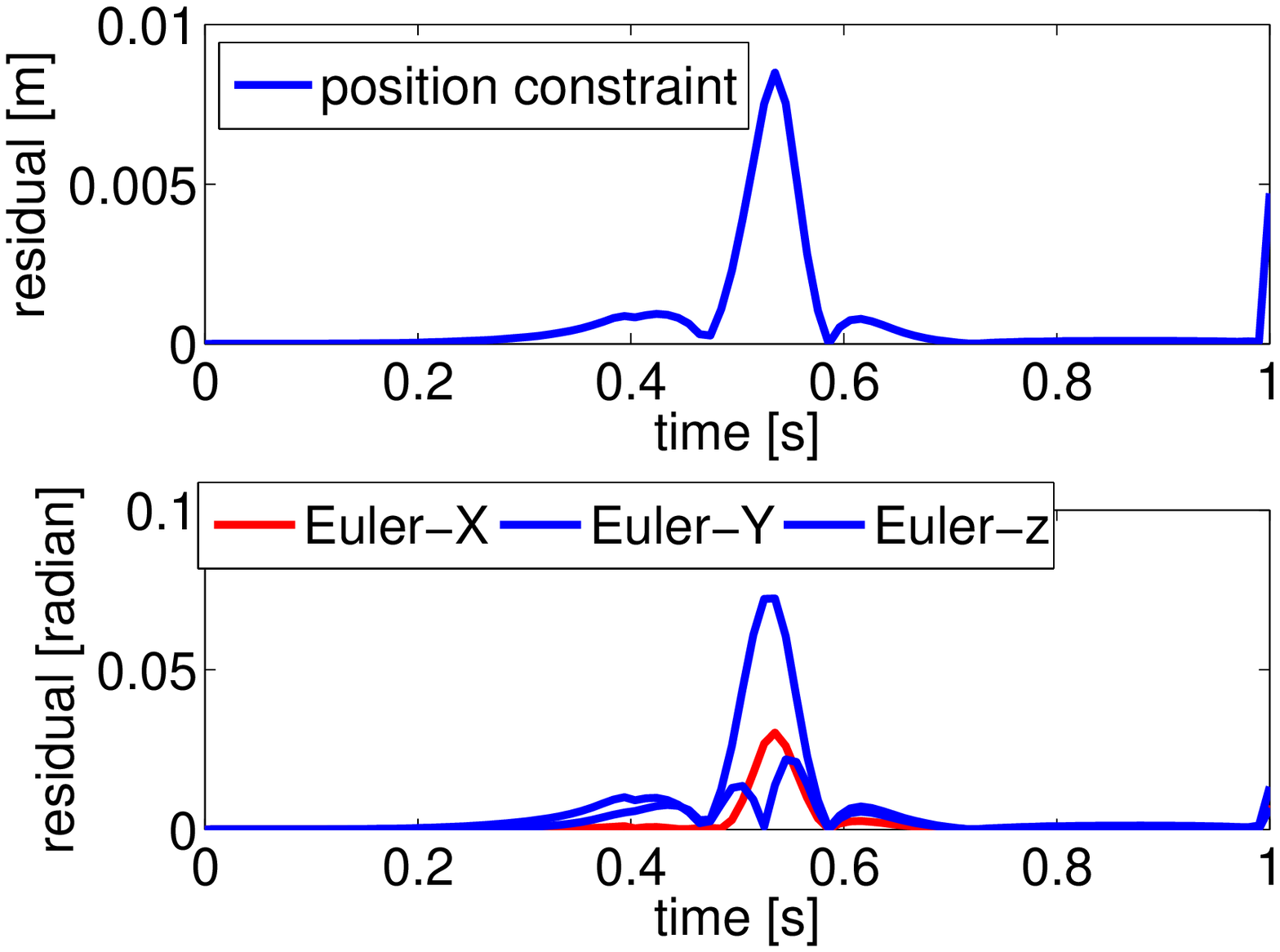}
    \label{kuka_path2_posres}
   }   
   \subfigure[]{
    \includegraphics[width= 4.35cm, height=3.6cm] {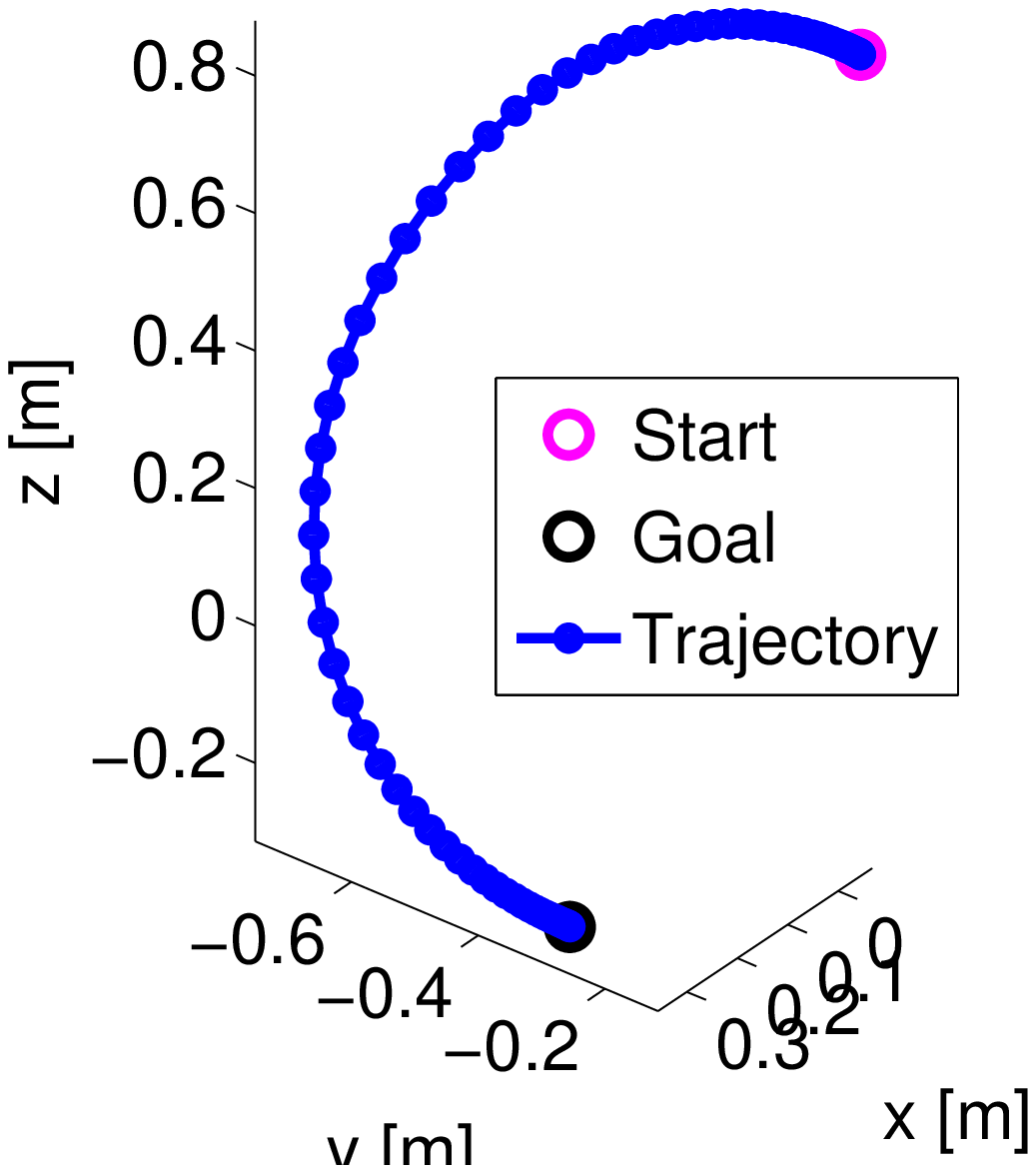}
    \label{kuka_path1}
   }\hspace{-0.7cm}
   \subfigure[]{
    \includegraphics[width= 4.35cm, height=3.6cm] {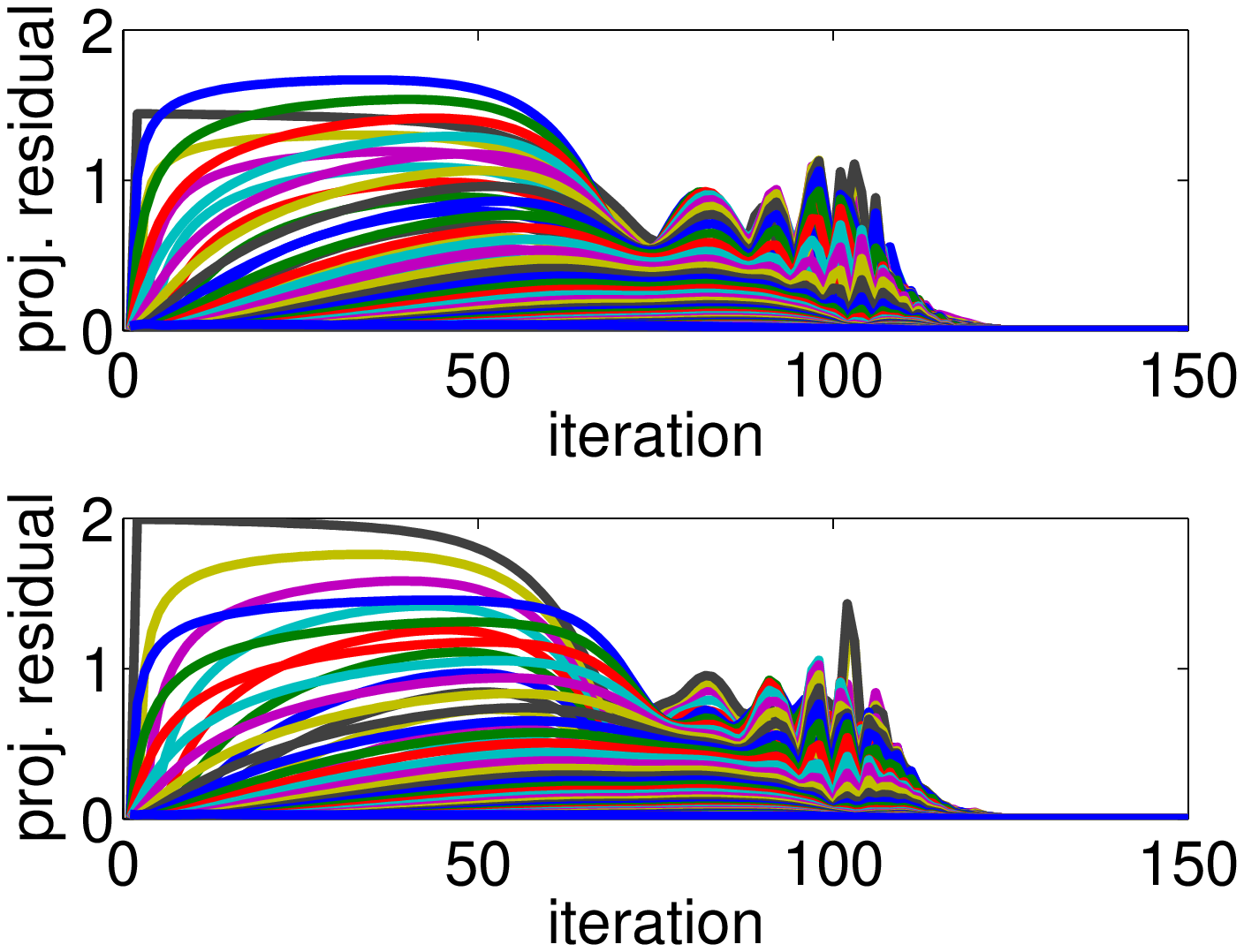}
    \label{kuka_path1_proj}
   }\hspace{-0.7cm}
   \subfigure[]{
      \includegraphics[width= 4.35cm, height=3.6cm] {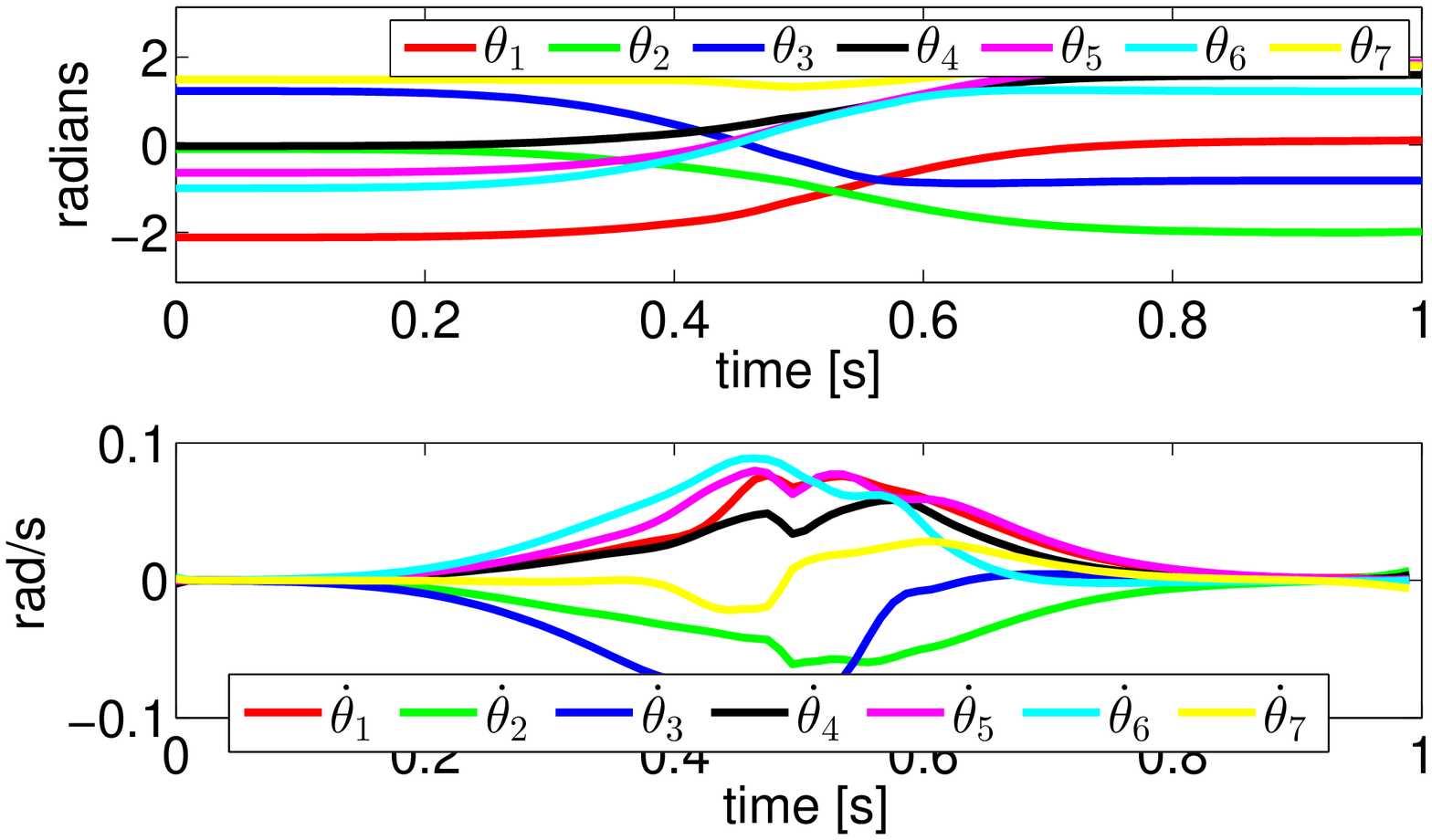}
    \label{kuka_path1_joint}
   }\hspace{-0.5cm}
   \subfigure[]{
      \includegraphics[width= 4.35cm, height=3.6cm] {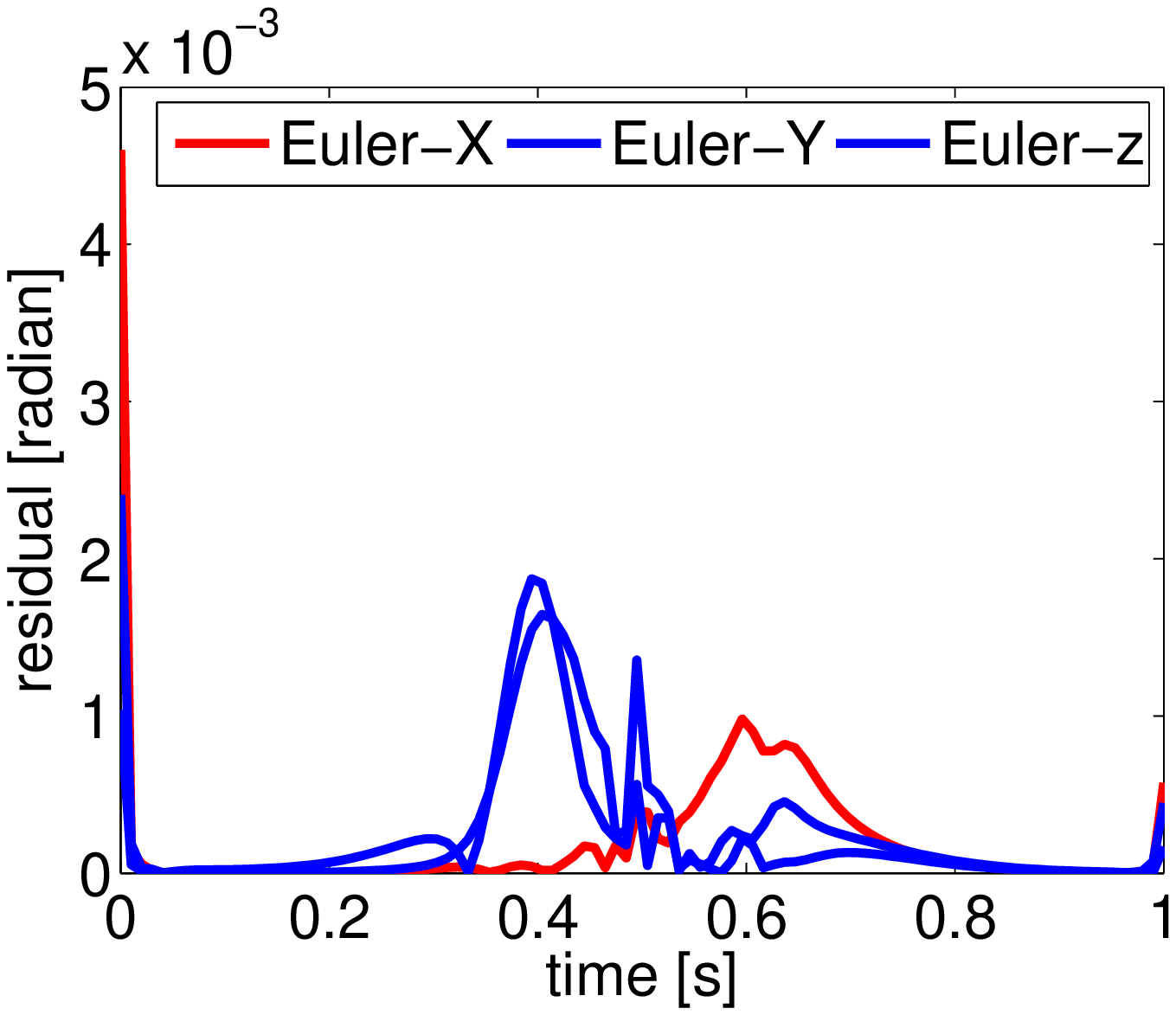}
    \label{kuka_path1_posres}
   }   
      
  \caption{(a) A cyclic task space trajectory executed by the KUKA LWR. (b) Residuals of (\ref{slack1})-(\ref{slack2}) with respect to iterations for the cyclic trajectory shown in (a). (c) Joint angles and velocities corresponding to the cyclic trajectory. (d) Task constraint residual for the cyclic trajectory. (e) Task space trajectory for point to point motion of KUKA LWR with trajectory wide constraints on $X$ component of position and orientation of the end effector. (f) residuals of (\ref{slack1})-(\ref{slack2}) with respect to iterations for the trajectory shown in (e). (g) Joint angles and velocity for trajectory shown in (e). (h) Task constraint residuals for the application shown in (e). (i) Task space trajectory for point to point motion with trajectory wide  constraints on end effector orientation. (j) Residuals of (\ref{slack1})-(\ref{slack2}) with respect to iterations for the trajectory shown in (i). (h) Joint angles and velocities for trajectory shown in (i). (h) Task constraint residuals for trajectory shown in (i). }          
  \vspace{-0.7cm}
\end{figure*}

In this application, a KUKA LWR manipulator is required to move between a given start and a goal position while keeping the orientation of the end effector fixed at $\textbf{R}_{XYZ}( 1.04,    0.0,  0.0 ) rad.$. The task constraint consists of nine equality constraints for the nine elements of the rotation matrix and three equality constraints for the final position. The results are summarized in Fig.\ref{kuka_path1}-\ref{kuka_path2_posres}. Note the smoothness in the joint angles and their velocities.

\begin{figure}
\centering
   \subfigure[]{
    \includegraphics[width= 4.35cm, height=3.4cm] {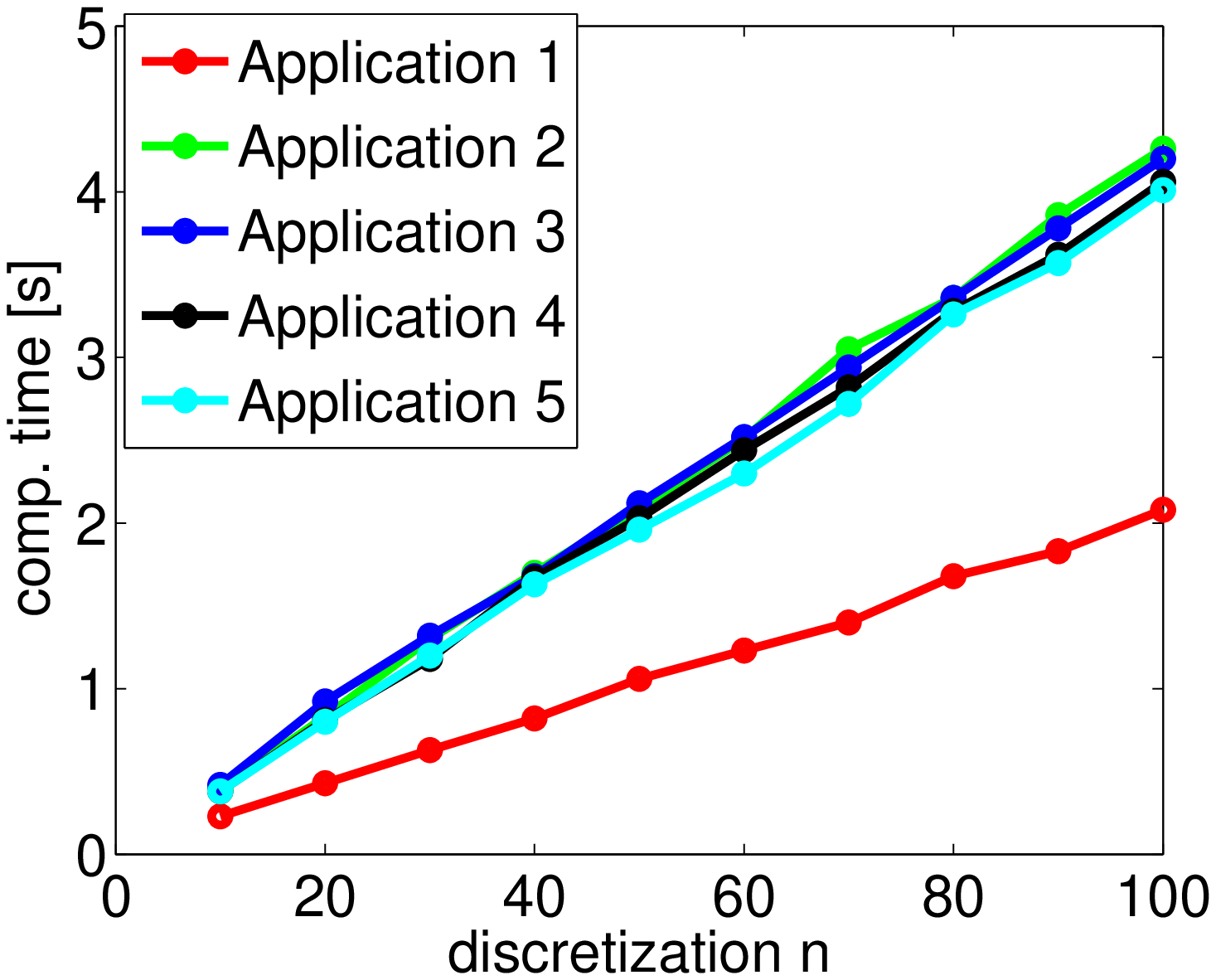}
    \label{comptime}
   }\hspace{-0.7cm}   
   \subfigure[]{
    \includegraphics[width= 4.35cm, height=3.4cm] {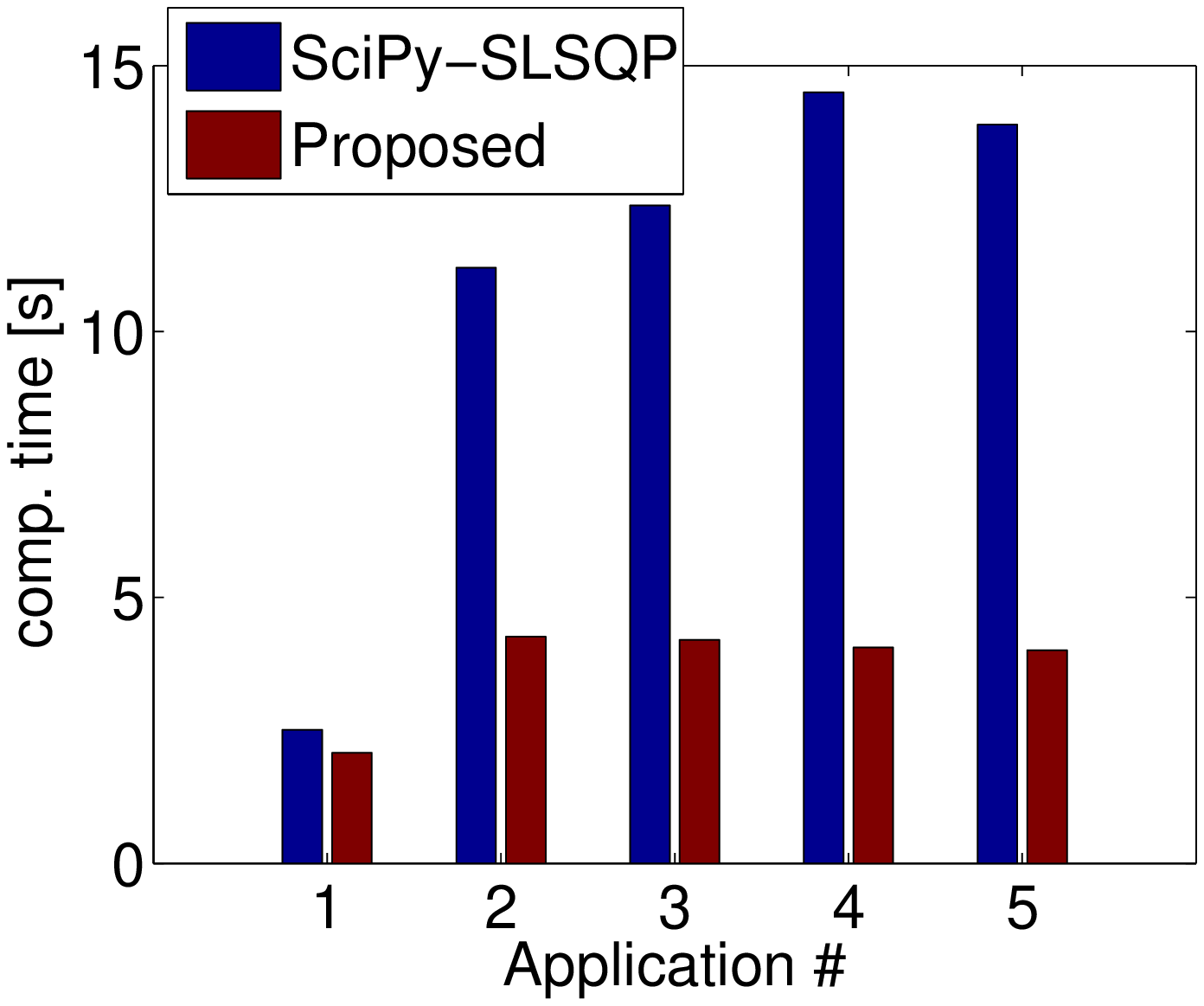}
    \label{comptime_compare}
   }\hspace{-0.7cm}
\caption{(a) computation time with respect to resolution of time discretization, $n$. (b) comparison of computation time between our proposed and SciPy-SLSQP for $n=100$.}
\end{figure}

\section{Computational Aspects}

Our entire implementation was done in Python using Numpy libraries on a laptop with 12GB RAM, $i7$ processor with $2.5Ghz$ clock speed. At the moment, our implementation does not exploit the massive parallelization opportunity provided by the proposed optimizer. In all the application problems discussed in the previous section, the size of the optimization problem was $m*n$, where $m$ is the number of joint angles and $n$ is the resolution of discretization of the time interval. Fig.\ref{comptime} shows the computational time for all the application problems as a function of $n$. Following points are noteworthy. Firstly, the computation time shows almost a linear growth with respect to $n$. Secondly, the computation time for the planar manipulator is significantly less than that obtained for KUKA LWR. In fact, the planning time for dual arm manipulator of application 2 which includes 12 $dofs$ is almost same as that obtained for  $7 dof$ planning of KUKA LWR. The reason for this could be traced back to the structure of the task constraints for a planar manipulator, which has the special form  of (\ref{nonlin_gen}). Once the reformulations (\ref{transform}) are used, the task constraints become affine. In contrast, the general form of task constraints, (\ref{nonlin_gen_spatial}) becomes affine for only a specific choice of the variables. Finally, the computation times are low enough to suggest that a combination of parallelization and prototyping in C can make the optimizer real time for  a small $n$ (typically, for $n=10$).

\noindent \textbf{Comparisons with an off-the-shelf optimizer} We compared our proposed optimizer with an off-the-shelf non-linear solver for a fixed resolution of discretization of $n=100$. Since our implementation is in Python, our comparison is with the optimizers provided in SciPy \cite{scipy}. In particular, with $SLSQP$ which implements the SQP based optimizer proposed in \cite{kraft}. For a fair comparison, we did not force $SLSQP$ to produce solutions with exceptionally low residuals and smoothness cost. Rather, we varied the iteration limit of $SLSQP$ to get a result comparable to our proposed optimizer in the quickest possible time. A comparison of computation times is presented in Fig.\ref{comptime_compare}. As can be seen, both our proposed and SLSQP optimizer have similar computation times for planar manipulators. However the difference increases sharply to upto more than three times on more complex applications with KUKA LWR or dual arm closed kinematic chain.

%
%
%

\section{Discussions and Future Work}
In this paper, we have shown  how the inherent structure prevalent in task constraints  can be exploited to significantly simplify the problem of task constrained trajectory optimization. We built our formulation on the very powerful yet simple concept of MAP which when combined with AL provided a globally valid convex approximation for the problem of task constrained trajectory optimization. We have demonstrated the usefulness of our formulation on some common benchmark problems and at the same shown it to be faster than off-the-shelf solvers provided in open source packages like SciPy.

Although, we have not incorporated collision avoidance in our formulation, we remark that its inclusion would not hamper the structure of the proposed convex approximation in any way. To see how, take the example of a planar manipulator. Any point on the manipulator would be an affine function of $\textbf{v}_t$, $\textbf{w}_t$ (see (\ref{nonlin_gen2}) and (\ref{planar_task1})-(\ref{planar_task2})). Thus, the euclidean distance between any point on the manipulator and a obstacle would have the same \emph{convex-concave} structure as that presented in  \cite{boyd_ccp2} (section 5.4). Convex approximation of \emph{convex-concave} constraints are well known and for the specific case of euclidean distance constraints, it reduces to a simple affine form. The extension to spatial manipulators can be done similarly following the discussion around (\ref{nonlin_gen_spatial2}).

There are various directions to extend our formulation. Of these the most interesting to our research are the following. (i) Evaluating whether the proposed formulation can be used to simplify the non-linear torque constraints and consequently contributing to the problem of task space inverse dynamics. (ii) Including obstacle avoidance within the optimizer. (iii) Cooperative transportation of objects. 

%

\bibliographystyle{IEEEtran}  
\bibliography{ref}

\begin{thebibliography}{10}
\providecommand{\url}[1]{#1}
\csname url@samestyle\endcsname
\providecommand{\newblock}{\relax}
\providecommand{\bibinfo}[2]{#2}
\providecommand{\BIBentrySTDinterwordspacing}{\spaceskip=0pt\relax}
\providecommand{\BIBentryALTinterwordstretchfactor}{4}
\providecommand{\BIBentryALTinterwordspacing}{\spaceskip=\fontdimen2\font plus
\BIBentryALTinterwordstretchfactor\fontdimen3\font minus
  \fontdimen4\font\relax}
\providecommand{\BIBforeignlanguage}[2]{{%
\expandafter\ifx\csname l@#1\endcsname\relax
\typeout{** WARNING: IEEEtran.bst: No hyphenation pattern has been}%
\typeout{** loaded for the language `#1'. Using the pattern for}%
\typeout{** the default language instead.}%
\else
\language=\csname l@#1\endcsname
\fi
#2}}
\providecommand{\BIBdecl}{\relax}
\BIBdecl

\bibitem{berenson}
D.~Berenson, S.~Srinivasa, and J.~Kuffner, ``Task space regions: A framework
  for pose-constrained manipulation planning,'' \emph{The International Journal
  of Robotics Research}, vol.~30, no.~12, pp. 1435--1460, 2011.

\bibitem{kino_manifold}
R.~Bordalba, L.~Ros, and J.~M. Porta, ``Kinodynamic planning on constraint
  manifolds,'' \emph{arXiv preprint arXiv:1705.07637}, 2017.

\bibitem{chomp}
M.~Zucker, N.~Ratliff, A.~D. Dragan, M.~Pivtoraiko, M.~Klingensmith, C.~M.
  Dellin, J.~A. Bagnell, and S.~S. Srinivasa, ``Chomp: Covariant hamiltonian
  optimization for motion planning,'' \emph{The International Journal of
  Robotics Research}, vol.~32, no. 9-10, pp. 1164--1193, 2013.

\bibitem{trajopt}
J.~Schulman, Y.~Duan, J.~Ho, A.~Lee, I.~Awwal, H.~Bradlow, J.~Pan, S.~Patil,
  K.~Goldberg, and P.~Abbeel, ``Motion planning with sequential convex
  optimization and convex collision checking,'' \emph{The International Journal
  of Robotics Research}, vol.~33, no.~9, pp. 1251--1270, 2014.

\bibitem{ccp1}
A.~L. Yuille and A.~Rangarajan, ``The concave-convex procedure,'' \emph{Neural
  computation}, vol.~15, no.~4, pp. 915--936, 2003.

\bibitem{ccp_robot1}
F.~Gao and S.~Shen, ``Quadrotor trajectory generation in dynamic environments
  using semi-definite relaxation on nonconvex qcqp,'' in \emph{Robotics and
  Automation (ICRA), 2017 IEEE International Conference on}.\hskip 1em plus
  0.5em minus 0.4em\relax IEEE, 2017, pp. 6354--6361.

\bibitem{ccp_robot2}
A.~K. Singh and K.~M. Krishna, ``A class of non-linear time scaling functions
  for smooth time optimal control along specified paths,'' in \emph{Intelligent
  Robots and Systems (IROS), 2015 IEEE/RSJ International Conference on}.\hskip
  1em plus 0.5em minus 0.4em\relax IEEE, 2015, pp. 5809--5816.

\bibitem{map1}
J.~von Neumann, ``Functional operators. vol. ii. the geometry of orthogonal
  spaces, volume 22 (reprint of 1933 notes) of annals of math,'' \emph{Studies.
  Princeton University Press}, 1950.

\bibitem{map2}
H.~H. Bauschke and J.~M. Borwein, ``On projection algorithms for solving convex
  feasibility problems,'' \emph{SIAM review}, vol.~38, no.~3, pp. 367--426,
  1996.

\bibitem{map3}
Z.~Zhu and X.~Li, ``Convergence analysis of alternating nonconvex
  projections,'' \emph{arXiv preprint arXiv:1802.03889}, 2018.

\bibitem{map4}
A.~S. Lewis, D.~R. Luke, and J.~Malick, ``Local linear convergence for
  alternating and averaged nonconvex projections,'' \emph{Foundations of
  Computational Mathematics}, vol.~9, no.~4, pp. 485--513, 2009.

\bibitem{al1}
M.~Powell, ``A method for nonlinear constraints in minimization problems in
  optimization ed. by r,'' 1969.

\bibitem{al2}
M.~R. Hestenes, ``Multiplier and gradient methods,'' \emph{Journal of
  optimization theory and applications}, vol.~4, no.~5, pp. 303--320, 1969.

\bibitem{kuka_cyclic}
M.~Cefalo, G.~Oriolo, and M.~Vendittelli, ``Planning safe cyclic motions under
  repetitive task constraints,'' in \emph{Robotics and Automation (ICRA), 2013
  IEEE International Conference on}.\hskip 1em plus 0.5em minus 0.4em\relax
  IEEE, 2013, pp. 3807--3812.

\bibitem{boyd_scp}
S.~Boyd, ``Sequential convex programming,'' \emph{Lecture Notes, Stanford
  University}, 2008.

\bibitem{toussiant}
M.~Toussaint, ``A tutorial on newton methods for constrained trajectory
  optimization and relations to slam, gaussian process smoothing, optimal
  control, and probabilistic inference,'' in \emph{Geometric and Numerical
  Foundations of Movements}.\hskip 1em plus 0.5em minus 0.4em\relax Springer,
  2017, pp. 361--392.

\bibitem{cuong_topp}
Q.-C. Pham, ``A general, fast, and robust implementation of the time-optimal
  path parameterization algorithm,'' \emph{IEEE Transactions on Robotics},
  vol.~30, no.~6, pp. 1533--1540, 2014.

\bibitem{aks_topp}
A.~K. Singh and K.~M. Krishna, ``A class of non-linear time scaling functions
  for smooth time optimal control along specified paths,'' in \emph{Intelligent
  Robots and Systems (IROS), 2015 IEEE/RSJ International Conference on}.\hskip
  1em plus 0.5em minus 0.4em\relax IEEE, 2015, pp. 5809--5816.

\bibitem{kuka}
R.~Bischoff, J.~Kurth, G.~Schreiber, R.~Koeppe, A.~Albu-Sch{\"a}ffer, A.~Beyer,
  O.~Eiberger, S.~Haddadin, A.~Stemmer, G.~Grunwald \emph{et~al.}, ``The
  kuka-dlr lightweight robot arm-a new reference platform for robotics research
  and manufacturing,'' in \emph{Robotics (ISR), 2010 41st international
  symposium on and 2010 6th German conference on robotics (ROBOTIK)}.\hskip 1em
  plus 0.5em minus 0.4em\relax VDE, 2010, pp. 1--8.

\bibitem{scipy}
E.~Jones, T.~Oliphant, and P.~Peterson, ``$\{$SciPy$\}$: open source scientific
  tools for $\{$Python$\}$,'' 2014.

\bibitem{kraft}
D.~Kraft, ``A software package for sequential quadratic programming,''
  \emph{Forschungsbericht- Deutsche Forschungs- und Versuchsanstalt fur Luft-
  und Raumfahrt}, 1988.

\bibitem{boyd_ccp2}
X.~Shen, S.~Diamond, Y.~Gu, and S.~Boyd, ``Disciplined convex-concave
  programming,'' \emph{arXiv preprint arXiv:1604.02639}, 2016.

\end{thebibliography}

\end{document}